%% file: Structure-preservingNN-ARXIV.tex
\documentclass[10pt,A4paper]{article}
\usepackage[T1]{fontenc}
\usepackage[utf8]{inputenc}
\usepackage{authblk}
\usepackage{arxiv}

\usepackage{amsmath,amssymb,amsfonts}\usepackage[colorlinks,bookmarksnumbered,bookmarksopen,citecolor=blue,urlcolor=blue,linkcolor=blue,]{hyperref}
\newcommand{\bs}[1]{\boldsymbol{#1}}
\usepackage{booktabs}

\newcommand{\mysecref}[1]{Section~\ref{#1}}
\newcommand{\myeqref}[1]{Eq.~(\ref{#1})}
\newcommand{\myfigref}[1]{Fig.~\ref{#1}}
\newcommand{\myalgref}[1]{Algorithm~\ref{#1}}
\newcommand{\mytabref}[1]{Table~\ref{#1}}

\newcommand{\dpar}[2]{\frac{\partial #1}{\partial #2}}
\newcommand{\diff}[2]{\frac{d #1}{d #2}}
\newcommand{\ddiff}[2]{\frac{\mathsf{D} #1}{\mathsf{D} #2}}

\usepackage{algorithm,algpseudocode}
\usepackage{algorithmicx}
\usepackage{caption} 
\usepackage{tikz}
\usepackage{subcaption}
\usepackage{multirow}
\usepackage{pgfplots}
\usetikzlibrary{intersections,shapes.arrows}
\usetikzlibrary{patterns, positioning, arrows,spy,shapes.geometric,calc}
\usetikzlibrary{shapes,arrows}

\usepackage{pgfplotstable}
\usepgfplotslibrary{statistics}
\pgfplotsset{
/pgfplots/custom legend/.style={
legend image code/.code={
\draw [only marks,mark=square]
plot coordinates {(0.3cm,0cm)};
}, },}

\title{Structure-preserving neural networks}

\author[1]{Quercus Hern\'andez}
\author[1]{Alberto Bad\'ias}
\author[1]{David Gonz\'alez}
\author[2]{Francisco Chinesta}
\author[1]{El\'ias Cueto}
\affil[1]{{\small Aragon Institute of Engineering Research. Universidad de Zaragoza. Zaragoza, Spain.}}
\affil[2]{{\small ESI Group chair. PIMM Lab. ENSAM Institute of Technology. Paris, France.}}

\begin{document}

\maketitle

\begin{abstract}
We develop a method to learn physical systems from data that employs feedforward neural networks and whose predictions comply with the first and second principles of thermodynamics. The method employs a minimum amount of data by enforcing the metriplectic structure of  dissipative Hamiltonian systems in the form of the so-called General Equation for the Non-Equilibrium Reversible-Irreversible Coupling, GENERIC [{\it M. Grmela and H.C Oettinger (1997). Dynamics and thermodynamics of complex fluids. I. Development of a general formalism. Phys. Rev. E. 56 (6): 6620--6632}]. The method does not need to enforce any kind of balance equation, and thus no previous knowledge on the nature of the system is needed. Conservation of energy and dissipation of entropy in the prediction of previously unseen situations arise as a natural by-product of the structure of the method. Examples of the performance of the method are shown that comprise conservative as well as dissipative systems, discrete as well as continuous ones.
\end{abstract}

\section{Introduction}
\label{Sec::Introduction}

With the irruption of the so-called fourth paradigm of science \cite{fourth-paradigm} a growing interest is detected on the machine learning of scientific laws. A plethora of methods have been developed that are able to produce more or less accurate predictions about the response of physical systems in previously unseen situations by employing techniques ranging from classical regression to the most sophisticated deep learning methods. 

For instance, recent works in solid mechanics have  substituted the constitutive equations with experimental  data \cite{kirchdoerfer2016data,ayensa2018datadriven}, while conserving the traditional approach on physical laws with high epistemic value (i.e., balance equations, equilibrium). Similar approaches have applied this concept to the unveiling (or correction) of plasticity models \cite{ibanez2019hybrid}, while others  created the new concept of constitutive manifold \cite{ibanez2017data,ibanez2018manifold}. Other approaches are designed to unveil an explicit, closed form expression for the physical law governing the phenomenon at hand \cite{Brunton}.

An interest is observed in the incorporation of the already existing scientific knowledge to these data-driven procedures. This interest is two-fold. Indeed, we prefer not to get rid of centuries of scientific knowledge and rely exclusively on powerful machine learning strategies. Existing theories have proved to be useful in the prediction of physical phenomena and are still in the position of helping to produce very accurate predictions.  This is the procedure followed in the so-called data-driven computational mechanics approach mentioned before. On the other hand, these theories help to keep the consumption of data to a minimum. Data are expensive to produce and to maintain. Already existing scientific knowledge could alleviate the amount of data needed to produce a successful prediction.

The mentioned works on data-driven computational mechanics usually rely on traditional machine learning algorithms, which are very precise and tested but usually computationally expensive. With the recent advances in data processing, computing resources and machine learning, neural networks have become a powerful tool to analyze traditionally hard problems such as image classification \cite{krizhevsky2012imagenet, shin2016deep}, speech recognition \cite{hinton2012deep, graves2013speech} or data compressing \cite{theis2017lossy, romero2017quantum}. These new machine learning methods outperform many of the traditional ones, both in modelling capacity and computational time (once trained, certain neural networks can easily handle real time requirements). Recent work in the machine learning community \cite{marquez2017imposing,lee2019gradient,nandwani2019primal} have shown that neural networks are also versatile in constraint optimizations.

This is the approach followed by several authors in the context of physical simulations, which aim to solve a set of partial differential equations (PDEs) in complex dynamical systems. Physical problems must satisfy inherently certain conditions dictated by physics, often formulated as conservation laws, and can be imposed to a neural network using extra loss terms in the constrained optimization process \cite{magiera2020constraint}.

Similar constraints are imposed in the so-called physically-informed neural networks approach \cite{raissi2019physics, zhang2020learning}. This family of methods employs neural networks to solve highly nonlinear partial differential equations (PDEs) resulting in very accurate and numerically stable results. However, they rely on prior knowledge of the governing equations of the problem. 

The authors have introduced the so-called thermodynamically consistent data-driven computational mechanics \cite{gonzalez2018datadriven,gonzalez2019hyperelastic,ghnatios2019poroviscolastic}. Unlike other existing works, this approach does not impose any particular balance equation to solve for. Instead, it relies on the imposition of the right thermodynamic structure of the resulting predictions, as dictated by the so-called GENERIC formalism \cite{grmela1997dynamics}. As will be seen, this ensures conservation of energy and the right amount of entropy dissipation, thus giving rise to predictions satisfying the first and second principles of thermodynamics. These techniques, however, employ regression to unveil the thermodynamic structure of the problem at the sampling points. For previously unseen situations, they employ interpolation on the matrix manifold describing the system.

Recent work in symplectic networks \cite{jin2020symplectic} have by-passed those drawbacks by exploiting the mathematical properties of Hamiltonian systems, so no prior knowledge of the system is required. However, this technique only operates on conservative systems with no entropy generation.

The aim of this work is the development of a new structure-preserving neural network architecture capable of predicting the time evolution of a system based on experimental observations on the system, with no prior knowledge of its governing equations, to be valid for both conservative and dissipative systems. The key idea is to merge the proven computational power of neural networks in highly nonlinear physics with thermodynamic consistent data-driven algorithms. The resulting methodology, as will be seen, is a powerful neural network architecture, conceptually very simple---based on standard feedforward methodologies---that exploits the right thermodynamic structure of the system as unveiled from experimental data, and that produces interpretable results \cite{murdoch2019interpretable}.

The outline of the paper is as follows. A brief description of the problem setup is presented in \mysecref{sec:prob}. Next, in \mysecref{sec:method}, the methodology is presented of both the GENERIC formalism and the feed-forward neural networks used to solve the stated problem. This technique is used in different physical systems of increasing complexity: a double thermo-elastic pendulum (\mysecref{sec:double_pendulum}) and a Couette flow in a viscoelastic fluid (\mysecref{sec:visco}). The paper is completed with a discussion in \mysecref{sec:conc}. 

\section{Problem Statement}\label{sec:prob}

Weinan E seems to be the first author in interpreting the process of learning physical systems as the solution of a dynamical system \cite{E2017}. Consider a system whose governing variables will be hereafter denoted by $\bs z \in \mathcal M \subseteq \mathbb R^n$, with $\mathcal M$ the state space of these variables, which is assumed to have the structure of a differentiable manifold in $\mathbb R^n$.

The problem of learning a given physical phenomenon can thus be seen as the one of finding an expression for the time evolution of their governing variables $\bs z$,
\begin{equation}\label{eq:pde}
\dot{\bs z}= \diff{\bs{z}}{t} = F(\bs{x},\bs{z},t),\; \bs{x}\in\Omega\in\mathbb{R}^{D},\; t\in\mathcal{I}=(0,T],\; \bs z(0)=\bs z_0,
\end{equation}
where $\bs{x}$ and $t$ refer to the space and time coordinates within a domain with $D=2,3$ dimensions. $F(\bs{x},\bs z, t)$ is the function that gives, after a prescribed time horizon $T$, the flow map $\bs z_0 \rightarrow \bs z(\bs z_0,T)$.

While this problem can be seen as a general supervised learning problem (we fix both $\bs z_0$ and $\bs z$), when we have additional information about the physics being represented by the sought function $F$, it is legitimate to try to include it in the search procedure. W. E seems to have been the first in suggesting to impose a Hamiltonian structure on $F$ if we know that energy is conserved, for instance \cite{E2017}. Very recently, two different approaches follow this same rationale \cite{KevrekidisHamiltonian,jin2020symplectic}.

For conservative systems, therefore, imposing a Hamiltonian structure seems a very appealing way to obtain thermodynamics-aware results. However, when the system is dissipative, this method does not provide with valid results. Given the importante of dissipative phenomena (viscous solids, fluid dynamics, ...) we explore the right thermodynamic structure to impose to the search methodology.

The goal of this paper is to develop a new method of solving \myeqref{eq:pde} using state of the art deep learning tools, in order to predict the time evolution of the state variables of a given system. The solution is forced to fulfill the basic thermodynamic requirements of energy conservation and entropy inequality restrictions via the GENERIC formalism, presented in the next section.

\section{Methodology}\label{sec:method}

In this section we develop the appropriate thermodynamic structure for dissipative systems. Classical systems modeling can be done at a variety of scales. We could think of the most detailed (yet often impractical) scale of molecular dynamics, where energy conservation applies and the Hamiltonian paradigm can be imposed. However, the number of degrees of freedom and, noteworthy, the time scale, renders this approach of little interest for many applications. On the other side of the spectrum lies thermodynamics, where only conserved, invariant, quantities are described and thus there is no need for conservation principles. At any other (mesoscopic) scale, unresolved degrees of freedom give rise to the appearance of fluctuation in the results (or its equivalent, dissipation). At these scales, traditional modeling procedures imply expressing physical insights in the form of governing equations \cite{grmela2018generic}. These equations are then validated from experimental observations. 

Alternatively, thermodynamics can be thought of as  a {\it meta-physics}, in the sense that it is actually a theory of theories \cite{grmela2019gradient}. It provides us with the right theoretic framework in which basic principles are met. And, in particular for any of these intermediate or mesoscopic scales, a so-called metriplectic structure emerges. The term metriplectic comes for the combination of symplectic and Riemannian (metric)  geometry and emphasizes the fact that there are conservative as well as dissipative contributions to the general evolution of such a system. Once such a geometric structure is found for the system, we are in the position of fixing the framework in which our neural networks can look for the adequate prediction of the future states of the system. The particular metriplectic structure that we employ for such a task is known, as stated before, as GENERIC.

\subsection{The GENERIC Formalism}

The ``General Equation for Non-Equilibrium Reversible-Irreversible Coupling'', GENERIC, formalism \cite{grmela1997dynamics,ottinger1997dynamics} establishes a mathematical framework in order to model the dynamics of a system. Furthermore, it is compatible with classical equilibrium thermodynamics \cite{ottinger2005beyond}, preserving the symmetries of the system as stated in Noether's theorem. It has served as the basis for the development of several consistent numerical integration algorithms that exploit these desirable properties \cite{romero2009thermodynamically, gonzalez2019thermodynamically}.

The GENERIC structure for the evolution in \myeqref{eq:pde} is obtained after finding two algebraic or differential operators 
$$
\bs L : T^*\mathcal M \rightarrow T\mathcal M,\quad \bs M : T^*\mathcal M \rightarrow T\mathcal M,
$$
where $T^*\mathcal M$ and $T\mathcal M$ represent, respectively, the cotangent and tangent bundles of $\mathcal M$. As in general Hamiltonian systems, there will be an energy potential, which we will denote hereafter by $E(\bs z)$. In order to take into account the dissipative effects, a second potential (the so-called Massieu potential) is introduced in the formulation. It is, of course, the entropy potential of the GENERIC formulation, $S(\bs z)$. With all these ingredients, we arrive at a description of the dynamics of the system of the type
\begin{equation}\label{eq:generic}
 \diff{\bs{z}}{t} = \bs{L} \dpar{E}{\bs{z}} + \bs{M} \dpar{S}{\bs{z}}.
\end{equation}

As shown in \myeqref{eq:generic}, the time evolution of the system described by the nonlinear operator $F(\bs{x},\bs z, t)$ presented in \myeqref{eq:pde} is now split in two separated terms:
\begin{itemize}
\item \textbf{Reversible Term}: It accounts for all the reversible (non-dissipative) phenomena of the system. In the context of classical mechanics, this term is equivalent to Hamilton's equations of motion that relates the particle position and momentum. The operator $\bs{L}(\bs{z})$ is the Poisson matrix---it defines a Poisson bracket---and is required to be skew-symmetric (a cosymplectic matrix). 

\item \textbf{Non-Reversible Term}: The rest of the non-reversible (dissipative) phenomena of the system are modeled here. The operator $\bs{M}(\bs{z})$ is the friction matrix and is required to be symmetric and positive semi-definite. 
\end{itemize}

The GENERIC formulation of the problem is completed with the following so-called degeneracy conditions
\begin{equation}\label{eq:degen}
 \bs{L} \dpar{S}{\bs{z}}=\bs{M} \dpar{E}{\bs{z}} = \bs 0.
\end{equation}

The first condition express esthe reversible nature of the $\bs{L}$ contribution to the dynamics whereas the second requirement expresses the conservation of the total energy by the $\bs{M}$ contribution. This means no other thing that the energy potential does not contribute to the production of entropy and, conversely, that the entropy functional does not contribute to reversible dynamics.  This mutual degeneracy requirement in addition to the already mentioned $\bs{L}$ and $\bs{M}$ matrix requirements ensure that:
\begin{equation*}\label{eq:Econs}
 \dpar{E}{t} = \dpar{E}{\bs{z}}\cdot\dpar{\bs{z}}{t}=\dpar{E}{\bs{z}}\left( \bs{L} \dpar{E}{\bs{z}} + \bs{M} \dpar{S}{\bs{z}}\right)=0,
\end{equation*}
which expresses the conservation of energy in an isolated system, also known as the first law of thermodynamics. Applying the same reasoning to the entropy $S$:
\begin{equation*}\label{eq:Scons}
 \dpar{S}{t} = \dpar{S}{\bs{z}}\cdot\dpar{\bs{z}}{t}=\dpar{S}{\bs{z}}\left( \bs{L} \dpar{E}{\bs{z}} + \bs{M} \dpar{S}{\bs{z}}\right)=\dpar{S}{\bs{z}}\bs{M}\dpar{S}{\bs{z}}\geq 0,
\end{equation*}
which guarantees the entropy inequality, this is, the second law of thermodynamics.

\subsection{Proposed Integration Algorithm}

Once the learning procedure is accomplished, our neural network is expected to integrate the system dynamics in time, given previously unseen initial conditions. In order to numerically solve the GENERIC equation, we formulate the discretized version of \myeqref{eq:generic} following previous works \cite{gonzalez2019thermodynamically}:
\begin{equation}\label{eq:generic_disc}
 \frac{\bs{z}_{n+1}-\bs{z}_{n}}{\Delta t} = \mathsf{L}\cdot\ddiff{E}{\bs{z}} + \mathsf{M}\cdot\ddiff{S}{\bs{z}}.
\end{equation}

The time derivative of the original equation is discretized with a forward Euler scheme in time increments $\Delta t$, where $\bs{z}_{n+1}=\bs{z}_{t+\Delta t}$. $\mathsf{L}$ and $\mathsf{M}$ are the discretized versions of the Poisson and friction matrices. Last, $\ddiff{E}{\bs{z}}$ and $\ddiff{S}{\bs{z}}$ represent the discrete gradients, which can be approximated in a finite element sense as:
\begin{equation*}\label{eq:grad_approx}
 \ddiff{E}{\bs{z}}\simeq \bs{A}\bs{z},\quad
 \ddiff{S}{\bs{z}}\simeq\bs{B}\bs{z},
\end{equation*}
where $\bs{A}$ and $\bs{B}$ represent the discrete matrix form of the gradient operators.

Finally, manipulating algebraically \myeqref{eq:generic_disc} with \myeqref{eq:grad_approx} and including the degeneracy conditions of \myeqref{eq:degen}, the proposed integration scheme for predicting the dynamics of a physical system is the following
\begin{equation}\label{eq:integration}
 \bs{z}_{n+1}=\bs{z}_{n} + \Delta t \left( \mathsf{L}\cdot\bs{A}\bs{z}_n + \mathsf{M}\cdot\bs{B}\bs{z}_n \right)
\end{equation}
subject to:
\begin{eqnarray*}\label{eq:degendisc}
 \mathsf{L}\cdot\bs{B}\bs{z}_n =\bs  0, \nonumber\\
 \mathsf{M}\cdot\bs{A}\bs{z}_n = \bs 0,
\end{eqnarray*}
ensuring the thermodynamical consistency of the resulting model.

To sum up, the main objective of this work is to compute the form of the $\bs{A}(\bs{z})$ and $\bs{B}(\bs{z})$ gradient operator matrices, subject to the degeneracy conditions, in order to integrate the initial system state variables $\bs{z}_0$ over certain time steps $\Delta t$ of the time interval $\mathcal{I}$. Usually, the form of matrices $\bs L$ and $\bs M$ is known in advance, given the vast literature in the field. If necessary, these terms can also be computed \cite{gonzalez2019thermodynamically}. %To achieve this, we implement the deep learning tools described in the next section.

\subsection{Feed-Forward Neural Networks}

In the introduction we already mentioned the intrinsic power of neural networks in many fields. The main reason under the fact that neural networks are able to learn and reproduce  such a variety of problems is that they are considered to be universal approximators \cite{cybenko1989approximation,hornik1989multilayer}, meaning that they are capable of approximating any measurable function to any desired degree of accuracy. The main limitation of this technique is the correct selection of the tuning parameters of the network, also called hyperparameters.

Another universal approximator are polynomials, as they can approximate any infinitely differentiable function as a Taylor power series expansion. The main difference is that neural networks rely on composition of functions rather than sum of power series:
\begin{equation}\label{eq:net_comp}
 \hat{\bs{y}}=(f^{[L]}\circ f^{[L-1]}\circ\,...\,\circ f^{[l]} \circ\,...\,\circ f^{[2]}\circ f^{[1]})(\bs{x}).
\end{equation}

\myeqref{eq:net_comp} shows that the desired output $\hat{\bs{y}}$ from a defined input $\bs{x}$ of a neural network is a composition of different functions $f^{[l]}$ as building blocks of the network in $L$ total layers. The challenge is to select the best combination of functions in the correct order such that it approximates the solution of the studied problem.

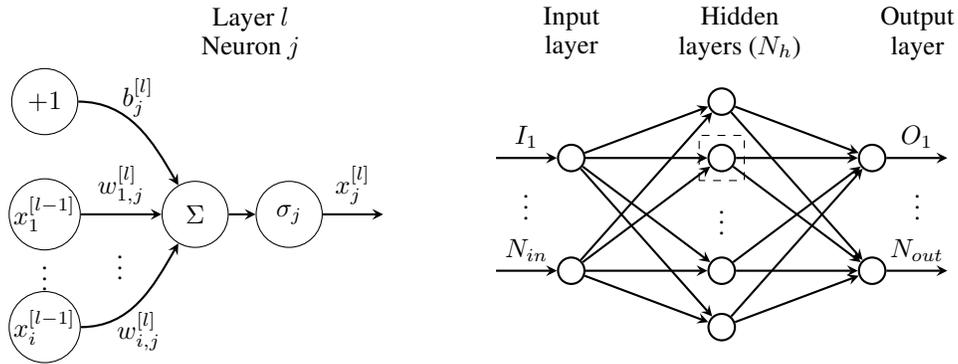
\begin{figure*}[h]
\centerline{
\input{./figures/neural_net/neural_net}
}
\caption{Representation of a single neuron (left) as a part of a fully connected neural net (right).\label{fig:fc_nn}}
\end{figure*}

The simplest building block of artificial deep neural network architectures is the neuron or perceptron (\myfigref{fig:fc_nn}, left). Several neurons are stacked in a multilayer perceptron (MLP), which is mathematically defined as follows
\begin{equation}\label{eq:forward_net}
 \bs{x}^{[l]} = \sigma(\bs{w}^{[l]}\bs{x}^{[l-1]}+\bs{b}^{[l]}),
\end{equation}
where $l$ is the index of the current layer, $\bs{x}^{[l-1]}$ and $\bs{x}^{[l]}$ are the layer input and output vector respectively, $\bs{w}^{[l]}$ is the weight matrix of the last layer, $\bs{b}^{[l]}$ is the bias vector of the last layer and $\sigma$ is the activation function. If no activation function is applied, the MLP is equivalent to a linear operator.  However, $\sigma$ is chosen to be a nonlinear function in order to increase the capacity of modelling more complex problems, which are commonly nonlinear. In classification problems, the traditional activation function is the logistic function (sigmoid) whereas in regression problems, Rectified Linear Unit (ReLU) \cite{glorot2011deep} or hyperbolic tangent are commonly used.

In this work, we use a deep neural network architecture known as feed-forward neural network \cite{schmidhuber2015deep}. It consists of a several layer of multilayer perceptrons with no cyclic connections, as shown in \myfigref{fig:fc_nn} (right).

The input of the neural net is the vector state of a given time step $\bs{z}_n$, and the outputs are the concatenated GENERIC matrices $\bs{A}_n^{\text{net}}$ and $\bs{B}_n^{\text{net}}$: for a system with $n$ state variables the number of inputs and outputs are $N_{\text{in}} = n$ and $N_{\text{out}}=2n^2$. Then, using the GENERIC integration scheme, the state vector at the next time step $\bs{z}_{n+1}^{\text{net}}$ is obtained. This method is repeated for the whole simulation time $T$ with a total of $N_T$ snapshots. 

The state variables of a general dynamical system may differ in several orders of magnitude from each other, due to their own physical nature or measurement units. Then, a pre-processing of the input data (scaling or normalization) can improve the model performance and stability.

The number of hidden layers $N_h$ depends on the complexity of the problem. Increasing the net size raises the computational power of the net to model more complex phenomena. However, it slows the training process and could lead to data overfitting, limiting its generalization and extrapolation capacity. The size of the hidden layers is chosen to be the same as the output size of the net $N_{\text{out}}$.

The cost function for our  neural network is composed of three different terms:
\begin{itemize}

\item \textbf{Data loss}: The main loss condition is the agreement between the network output and the real data. It is computed as the squared error sum, computed between the predicted state vector $\bs{z}_{n+1}^{\text{net}}$ and the ground truth solution $\bs{z}_{n+1}^{\text{GT}}$ for each time step. 
\begin{equation}\label{eq:cost_data}
\mathcal{L}_n^{\text{data}} = \Vert\bs{z}_{n+1}^{\text{GT}}-\bs{z}_{n+1}^{\text{net}}\Vert_2^2.
\end{equation}

\item \textbf{Fulfillment of the degeneracy conditions}: The cost function will also account for the degeneracy conditions in order to ensure the thermodynamic consistency of the solution, implemented as the sum of the squared elements of the degeneracy vectors for each time step,
\begin{equation}\label{eq:cost_degen}
\mathcal{L}_n^{\text{degen}} = \Vert\mathsf{L}\cdot\bs{B}_n^{\text{net}}\bs{z}_n^{\text{net}}\Vert_2^2 + \Vert\mathsf{M}\cdot\bs{A}_n^{\text{net}}\bs{z}_n^{\text{net}}\Vert_2^2.
\end{equation}
This term acts as a regularization of the loss function and, at the same time, is the responsible of ensuring thermodynamic consistency. So to speak, it is the cornerstone of our method.

\item \textbf{Regularization}: In order to avoid overfitting, an extra L2 regularization term $\mathcal{L}^{\text{reg}}$ is added to the loss function, defined as the sum over the squared weight parameters of the network.
\begin{equation}\label{eq:cost_regular}
\mathcal{L}^{\text{reg}}=\sum_l^{L}\sum_i^{n^{[l]}}\sum_j^{n^{[l+1]}}{(w_{i,j}^{[l]})^2}.
\end{equation}
\end{itemize}

The total cost function is computed as the sum squared error (SSE) of the data loss and degeneracy residual, in addition to the regularization term, at the end of the simulation time $T$ for each train case. The regularization loss is highly dependent on the size of the network layers and has different scaling with respect to the other terms, so it is compensated with the regularization hyperparameter (weight decay) $\lambda_r$. An additional weight $\lambda_d$ is added to the data loss term, which accounts for the relative scaling error with respect to the degeneracy conditions.

\begin{equation}\label{eq:cost_total}
\mathcal{L} = \sum_{n=0}^{N_T}{(\lambda_d \mathcal{L}_n^{\text{data}} + \mathcal{L}_n^{\text{degen}})} + \lambda_r \mathcal{L}^{\text{reg}}.
\end{equation}

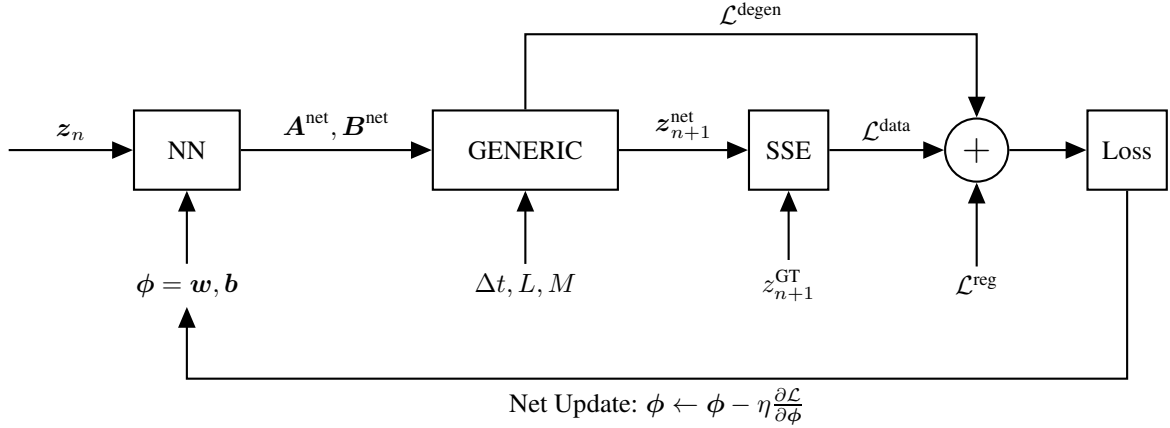
\begin{figure*}[h]
\centerline{
\input{./figures/algorithm_blocks/algorithm_blocks}
}
\caption{Sketch of a structure-preserving neural network training algorithm.\label{net}}
\end{figure*}

The usual backpropagation algorithm \cite{paszke2017automatic} is then used to calculate the gradient of the loss function for each net parameter (weight and bias vectors), which are updated with the gradient descent technique \cite{ruder2016overview}. The process is then repeated for a maximum number of epochs $n_{\text{epoch}}$. The resulting training algorithm is sketched in Fig. \ref{net}.

The proposed methodology is tested with two different databases of nonlinear physical systems, split in a partition of train cases ($N_{\text{train}}=80\%$ of the database) and test cases ($N_{\text{test}}=20\%$ of the database). The net performance is evaluated with the mean squared error (MSE) of the state variables prediction, associated with the data loss term, \myeqref{eq:cost_data}, over all the time snapshots,
\begin{equation}\label{eq:error_data}
\text{MSE}^{\text{data}}\;(\bs{z}_i)=\frac{1}{N_T}\sum_{n=0}^{N_T} \left(\bs{z}^{\text{GT}}_{i,n}-\bs{z}^{\text{net}}_{i,n}\right)^2.
\end{equation}
The same procedure is applied to the degeneracy constraint, associated with the degeneracy loss term, \myeqref{eq:cost_degen}, over all the time snapshots,
\begin{equation}\label{eq:error_degen}
\text{MSE}^{\text{degen}}\;(\bs{z}_i)=\frac{1}{N_T}\sum_{n=0}^{N_T} \left(\mathsf{L}\cdot\bs{B}_{i,n}^{\text{net}}\bs{z}_{i,n}^{\text{net}} + \mathsf{M}\cdot\bs{A}_{i,n}^{\text{net}}\bs{z}_{i,n}^{\text{net}}\right).
\end{equation}

\begin{algorithm*}[h!]
\caption{Pseudocode for the train algorithm.}\label{alg:net_train}
\begin{algorithmic}
  \State \textbf{Load train database:} $\bs{z}^{\text{GT}}$ (train partition), $\Delta t$, $\mathsf{L}$, $\mathsf{M}$;
  \State \textbf{Define network architecture:} $N_{\text{in}}$, $N_{\text{out}}=2N_{\text{in}}^2$, $N_{h}$, $\sigma_j$;
  \State \textbf{Define hyperparamteres:} $\eta$, $\lambda_d$, $\lambda_r$;
  \State Initialize $w_{i,j}$, $b_j$;
  \For{$epoch \gets 1,n_{\text{epoch}}$}
  	\For{$train\_case \gets 1,N_{\text{train}}$}
  		\State Initialize state vector: $\bs{z}_0^{\text{net}} \gets \bs{z}_0^{\text{GT}}$;
  		\State Initialize losses: $\mathcal{L}^{\text{data}},\mathcal{L}^{\text{degen}}=0$;
  		\For{$snapshot \gets 1,N_T$}
  			\State Forward propagation: $[\bs{A}_n^{\text{net}}$, $\bs{B}_n^{\text{net}}] \gets \text{Net}(\bs{z}_{n}^\text{GT})$; \Comment \myeqref{eq:forward_net}
  			\State Time integration: $\bs{z}_{n+1}^{\text{net}} \gets \bs{z}_{n}^{\text{net}} + \Delta t\;(\mathsf{L}\cdot\bs{A}_n^{\text{net}}\bs{z}_n^{\text{net}} + \mathsf{M}\cdot\bs{B}_n^{\text{net}}\bs{z}_{n}^{\text{net}})$; \Comment \myeqref{eq:generic_disc}
  			\State Update data loss: $\mathcal{L}^{\text{data}} \gets \mathcal{L}^{\text{data}} + \mathcal{L}_n^{\text{data}}$; \Comment \myeqref{eq:cost_data}
  			\State Update degeneracy loss: $\mathcal{L}^{\text{degen}} \gets \mathcal{L}^{\text{degen}} + \mathcal{L}_n^{\text{degen}}$; \Comment \myeqref{eq:cost_degen}
  		\EndFor		
  		\State SSE loss function: $L \gets \lambda_d \mathcal{L}^{\text{data}} + \mathcal{L}^{\text{degen}} + \lambda_r \mathcal{L}^{\text{reg}}$ \Comment \myeqref{eq:cost_regular}, \myeqref{eq:cost_total}
  		\State Backward propagation;
  		\State Optimizer step;
  	\EndFor
  \State Learning rate scheduler;
  \EndFor
\end{algorithmic}
\end{algorithm*}

As a general error magnitude of the algorithm, the average MSE of both the train ($N=N_{\text{train}}$) and test trajectories ($N=N_{\text{test}}$) is also reported for both the data ($m=\text{data}$) and degeneracy ($m=\text{degen}$) constraints,
\begin{equation}\label{eq:error_mean}
\overline{\text{MSE}}^m\;(\bs{z})=\frac{1}{N}\sum_{i=1}^{N}\text{MSE}^m\;(\bs{z}_i).
\end{equation}

\myalgref{alg:net_train} and \myalgref{alg:net_test} show a pseudocode of our proposed algorithm to both the training and test processes. The proposed method is fully implemented in PyTorch \cite{paszke2019pytorch} and trained in an Intel Core i7-8665U CPU.

\begin{algorithm*}[t]
\caption{Pseudocode for the test algorithm.}\label{alg:net_test}
\begin{algorithmic}
  \State \textbf{Load test database:} $\bs{z}^{\text{GT}}$ (test partition), $\Delta t$, $\mathsf{L}$, $\mathsf{M}$;
  \State \textbf{Load network parameters};
  	\For{$test\_case \gets 1,N_{test}$}
  		\State Initialize state vector: $\bs{z}_0^{\text{net}} \gets \bs{z}_0^{\text{GT}}$;
  		\For{$snapshot \gets 1,N_T$}
  			\State Forward propagation: $[\bs{A}_n^{\text{net}}$, $\bs{B}_n^{\text{net}}] \gets \text{Net}(\bs{z}_{n}^{\text{net}})$; \Comment \myeqref{eq:forward_net}
  			\State Time step integration: $\bs{z}_{n+1}^{\text{net}} \gets \bs{z}_{n}^{\text{net}} + \Delta t\;(\mathsf{L}\cdot\bs{A}_n^{\text{net}}\bs{z}_n^{\text{net}} + \mathsf{M}\cdot\bs{B}_n^{\text{net}}\bs{z}_{n}^{\text{net}})$; \Comment \myeqref{eq:generic_disc}
  			\State Update state vector: $\bs{z}_{n}^{\text{net}} \gets \bs{z}_{n+1}^{\text{net}}$;
  			\State Update snapshot: $n \gets n+1$;
  		\EndFor	
  		\State Compute $\text{MSE}^{\text{data}}$, $\text{MSE}^{\text{degen}}$; \Comment \myeqref{eq:error_data}, \myeqref{eq:error_degen}
  	\EndFor
  	\State Compute $\overline{\text{MSE}}^\text{data}$, $\overline{\text{MSE}}^{\text{degen}}$; \Comment \myeqref{eq:error_mean}
\end{algorithmic}
\end{algorithm*}

\newpage
\section{Validation examples: Double Thermo-Elastic Pendulum}\label{sec:double_pendulum}

\subsection{Description}

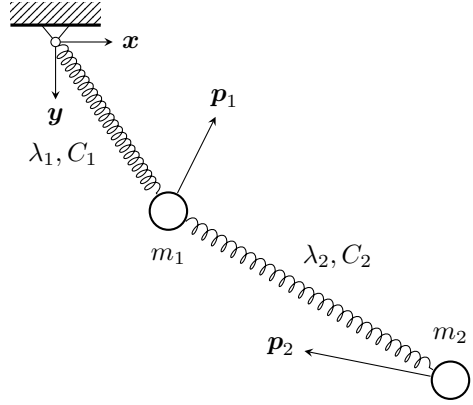
\begin{figure}[h]
\centerline{
\input{./figures/double_pendulum/double_pendulum}
}
\caption{Double thermo-elastic pendulum.\label{fig:double}}
\end{figure}

The first example is a double thermo-elastic pendulum (\myfigref{fig:double}) consisting of two masses $m_1$ and $m_2$ connected by two springs of variable lengths $\lambda_1$ and $\lambda_2$ and natural lengths at rest $\lambda_1^0$ and $\lambda_2^0$.

The set of variables describing the double pendulum are here chosen to be
\begin{equation}\label{eq:double_z}
\mathcal{S}=\{\bs{z}=(\bs{q}_1,\bs{q}_2,\bs{p}_1,\bs{p}_2,s_1,s_2)\in(\mathbb{R}^2\times\mathbb{R}^2\times\mathbb{R}^2\times\mathbb{R}^2\times\mathbb{R}\times\mathbb{R}),\quad \bs{q}_1\neq\bs{0}, \;\bs{q}_1\neq\bs{q}_2\}.
\end{equation}
where $\bs{q}_i$, $\bs{p}_i$ and $s_i$ are the position, linear momentum and entropy of each mass $i=1,2$.

The lengths of the springs $\lambda_1$ and $\lambda_2$ are defined solely in terms of the positions as
\begin{equation*}\label{eq:double_lambdas}
\lambda_1=\sqrt{\bs{q}_1\cdot\bs{q}_1},\quad \lambda_2=\sqrt{(\bs{q}_2-\bs{q}_1)\cdot(\bs{q}_2-\bs{q}_1}).
\end{equation*}

The total energy of the system can be expressed as the sum of the kinetic energy of the two masses $K_i$ and the internal energy of the springs $e_i$ for $i=1,2$,
\begin{alignat}{1}\label{eq:double_energy}
E&=E(\bs{z})=\sum_i K_i(\bs{z})+\sum_i e_i(\lambda_i,s_i),\nonumber\\
K_i&=\frac{1}{2m_i}|\bs{p}_i|^2.
\end{alignat}

The total entropy of the double pendulum is the sum of the entropies of the two masses $s_i$,
\begin{equation}\label{eq:double_entropy}
S=S(\bs{z})=s_1+s_2.
\end{equation}

This model includes thermal effects in the stretching of the springs due to the Gough-Joule effect. The absolute temperatures $T_i$ at each spring is obtained through \myeqref{eq:double_temperature}. These temperature changes induce a heat flux between both springs, being proportional to the temperature difference and a conductivity constant $\kappa>0$,
\begin{equation}\label{eq:double_temperature}
T_i=\dpar{e_i}{s_i}.
\end{equation}

In this case, there is a clear contribution of both conservative Hamiltonian mechanics (mass movement) and non-Hamiltonian dissipative effects (heat flux), resulting in a non-zero Poisson matrix ($\mathsf{M}\neq\bs{0}$). Thus, the GENERIC matrices associated with this physical system are known to be \cite{gonzalez2019thermodynamically}
\begin{equation}\label{eq:double_LM}
    \mathsf{L} = \begin{bmatrix} 
           		\bs{0} & \bs{0} & \bs{1} & \bs{0} & \bs{0} & \bs{0} \\
           		\bs{0} & \bs{0} & \bs{0} & \bs{1} & \bs{0} & \bs{0} \\
           		\bs{-1} & \bs{0} & \bs{0} & \bs{0} & \bs{0} & \bs{0} \\
           		\bs{0} & \bs{-1} & \bs{0} & \bs{0} & \bs{0} & \bs{0} \\
           		\bs{0} & \bs{0} & \bs{0} & \bs{0} & 0 & 0 \\
           		\bs{0} & \bs{0} & \bs{0} & \bs{0} & 0 & 0 \\
              \end{bmatrix},
\quad
    \mathsf{M} = \begin{bmatrix} 
           		\bs{0} & \bs{0} & \bs{0} & \bs{0} & \bs{0} & \bs{0} \\
           		\bs{0} & \bs{0} & \bs{0} & \bs{0} & \bs{0} & \bs{0} \\
           		\bs{0} & \bs{0} & \bs{0} & \bs{0} & \bs{0} & \bs{0} \\
           		\bs{0} & \bs{0} & \bs{0} & \bs{0} & \bs{0} & \bs{0} \\
           		\bs{0} & \bs{0} & \bs{0} & \bs{0} & 1 & -1/2 \\
           		\bs{0} & \bs{0} & \bs{0} & \bs{0} & -1/2 & 1 \\
              \end{bmatrix}.
\end{equation}

\subsection{Database and Hyperparameters}

The training database is generated with a thermodynamically consistent time-stepping algorithm\cite{romero2009thermodynamically} in MATLAB. The masses of the double pendulum are set to $m_1 = 1$ kg and $m_2=2$ kg, joint with springs of a natural length of $\lambda^0_1=2$ m and $\lambda^0_2=1$ m and thermal constant of $C_1=0.02$ J and $C_2=0.2$ J and conductivity constant of $\kappa = 0.5$. The simulation time of the movement is $T = 60$ s in time increments of $\Delta t = 0.3$ s ($N_T=200$ snapshots).

The database consists of the state vector, \myeqref{eq:double_z}, of 50 different trajectories with random initial conditions of position $\bs{q}_i$ and linear momentum $\bs{p}_i$ of both masses $m_i$ ($i=1,2$) around a mean position and linear momentum of $\bs{q}_1=[4.5,\;4.5]^\top$ m, $\bs{p}_1=[2,\;4.5]^\top$ kg$\cdot$m/s, and $\bs{q}_2=[-0.5,\;1.5]^\top$ m, $\bs{p}_2=[1.4,\;-0.2]^\top$ kg$\cdot$m/s respectively. Although the initial conditions of the simulations are similar, it results in a wide variety of the mass trajectories due to the chaotic behavior of the system. This database is split randomly in 40 train trajectories and 10 test trajectories. Thus, there is a total of $80.000$  training snapshots and $20.000$ test snapshots.

The net input and output size is $N_{\text{in}} = 10$ and $N_{\text{out}} = 2N_{\text{in}}^2 = 200$. The state vector is normalized based on the training set statistical mean and standard deviation. The number of hidden layers is $N_h = 5$ with ReLU activation functions and linear in the last layer. It is initialized according to the Kaiming method \cite{he2015delving} with normal distribution and the optimizer used is Adam \cite{kingma2014adam} with a weight decay of $\lambda_r=10^{-5}$ and data loss weight of $\lambda_d = 10^2$. A multistep learning rate scheduler is used, starting in $\eta=10^{-3}$ and decaying by a factor of $\gamma=0.1$ in epochs 600 and 1200. The training process ends when a fixed number of epochs $n_{epoch}=1800$ is reached.

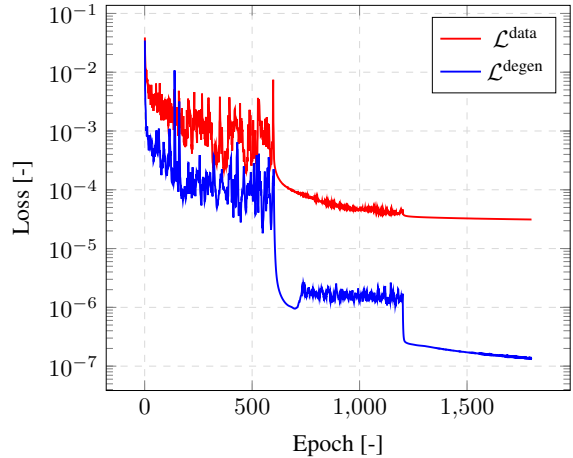
\begin{figure*}[h!]
\centering
\begin{tikzpicture}[scale=0.9]
\begin{semilogyaxis}[name = plot1,
  grid=major, % Display a grid
  grid style={dashed,gray!30}, % Set the style
  xlabel={$\text{Epoch}$ [-]},
  ylabel={$\text{Loss}$ [-]},
  legend pos=north east
  ]
  
	\addplot [red,thick] table [y=loss_error, x=epoch]{graphs/double_pendulum/double_log.txt};
	\addlegendentry{$\mathcal{L}^\text{data}$}
	\addplot [blue,thick] table [y=loss_degeneracy, x=epoch]{graphs/double_pendulum/double_log.txt};
	\addlegendentry{$\mathcal{L}^\text{degen}$}
\end{semilogyaxis}

\end{tikzpicture}
\caption{Loss evolution of data and degeneracy constraints for each epoch of the structure-preserving neural network training process of the double pendulum example.}
\label{fig:double_log}
\end{figure*}

The time evolution of the data $\mathcal{L}^\text{data}$ and degeneracy $\mathcal{L}^\text{degen}$ loss terms for each training epoch are shown in \myfigref{fig:double_log}.

\subsection{Results}

\begin{figure*}[h!]
\centering
\begin{tikzpicture}[scale=0.9]

%q1
\begin{axis}[name=plot1,
  grid=major, % Display a grid
  grid style={dashed,gray!30}, % Set the style
  xlabel={$t$ [s]},
  ylabel={$\bs{q}_1$ [m]},
  legend pos=south west
]
  
  \foreach \F in {6}{  %{1,2,...,10}{
	\addplot [color=blue, thick] table [y=qx1_net, x=tspan]{graphs/double_pendulum/double_results_test_\F.txt};
	\ifthenelse{\F=6}{\addlegendentry{SPNN (X)}}{}
	\addplot [color=red, thick] table [y=qy1_net, x=tspan]{graphs/double_pendulum/double_results_test_\F.txt};
	\ifthenelse{\F=6}{\addlegendentry{SPNN (Y)}}{}
	\addplot [color=black, thick, dashed, forget plot] table [y=qy1_real, x=tspan]{graphs/double_pendulum/double_results_test_\F.txt};
	\addplot [color=black, thick, dashed] table [y=qx1_real, x=tspan]{graphs/double_pendulum/double_results_test_\F.txt};
	\ifthenelse{\F=6}{\addlegendentry{GT}}{}
	}
\end{axis}

%q2    at={($(plot1.east)+(2cm,0)$)},
\begin{axis}[name=plot3, at={($(plot1.below south east)+(0,-0.5cm)$)}, anchor=above north east,
  grid=major, % Display a grid
  grid style={dashed,gray!30}, % Set the style
  xlabel={$t$ [s]},
  ylabel={$\bs{q}_2$ [m]},
  ]
  
  \foreach \F in {6}{  %{1,2,...,10}{
	\addplot [color=blue, thick] table [y=qx2_net, x=tspan]{graphs/double_pendulum/double_results_test_\F.txt};
	%\ifthenelse{\F=6}{\addlegendentry{SPNN (X)}}{}
	\addplot [color=black, thick, dashed] table [y=qx2_real, x=tspan]{graphs/double_pendulum/double_results_test_\F.txt};	
	\addplot [color=red, thick] table [y=qy2_net, x=tspan]{graphs/double_pendulum/double_results_test_\F.txt};
	%\ifthenelse{\F=6}{\addlegendentry{SPNN (Y)}}{}
	\addplot [color=black, thick, dashed, forget plot] table [y=qy2_real, x=tspan]{graphs/double_pendulum/double_results_test_\F.txt};
	%\ifthenelse{\F=6}{\addlegendentry{GT}}{}
	}
\end{axis}

%p2
\begin{axis}[name=plot4, at={($(plot3.right of north east)+(0.7cm,0)$)}, anchor=left of north west,
  grid=major, % Display a grid
  grid style={dashed,gray!30}, % Set the style
  xlabel={$t$ [s]},
  ylabel={$\bs{p}_2$ [kg·m/s]},
  ]
  
  \foreach \F in {6}{  %{1,2,...,10}{
	\addplot [color=blue, thick] table [y=px2_net, x=tspan]{graphs/double_pendulum/double_results_test_\F.txt};
	%\ifthenelse{\F=6}{\addlegendentry{SPNN (X)}}{}
	\addplot [color=black, thick, dashed] table [y=px2_real, x=tspan]{graphs/double_pendulum/double_results_test_\F.txt};
	%\ifthenelse{\F=6}{\addlegendentry{GT}}{}
	\addplot [color=red, thick] table [y=py2_net, x=tspan]{graphs/double_pendulum/double_results_test_\F.txt};
	%\ifthenelse{\F=6}{\addlegendentry{SPNN (Y)}}{}
	\addplot [color=black, thick, dashed, forget plot] table [y=py2_real, x=tspan]{graphs/double_pendulum/double_results_test_\F.txt};
	}
\end{axis}

%p1
\begin{axis}[name=plot2, at={($(plot4.above north west)+(0,0.5cm)$)}, anchor=below south west,
  grid=major, % Display a grid 
  grid style={dashed,gray!30}, % Set the style
  xlabel={$t$ [s]},
  ylabel={$\bs{p}_1$ [kg·m/s]},
  %legend style={at={(0.3,1.35)},anchor=west},
  %legend pos=outer north east
  ]
  
  \foreach \F in {6}{  %{1,2,...,10}{
	\addplot [color=blue, thick] table [y=px1_net, x=tspan]{graphs/double_pendulum/double_results_test_\F.txt};
	%\ifthenelse{\F=6}{\addlegendentry{SPNN (X)}}{}
	\addplot [color=black, thick, dashed] table [y=px1_real, x=tspan]{graphs/double_pendulum/double_results_test_\F.txt};
	%\ifthenelse{\F=6}{\addlegendentry{GT}}{}
	\addplot [color=red, thick] table [y=py1_net, x=tspan]{graphs/double_pendulum/double_results_test_\F.txt};
	%\ifthenelse{\F=6}{\addlegendentry{SPNN (Y)}}{}
	\addplot [color=black, thick, dashed, forget plot] table [y=py1_real, x=tspan]{graphs/double_pendulum/double_results_test_\F.txt};
	}
\end{axis}

\end{tikzpicture}
\caption{Time evolution of the state variables in a test trajectory of a double themo-elastic pendulum using a time-stepping solver (Ground Truth, GT) and the proposed GENERIC integration scheme (SPNN). Since every variable has a vectorial character, both components are depicted and labelled as $X$ and $Y$, respectively.}
\label{fig:double_results_time}
\end{figure*}
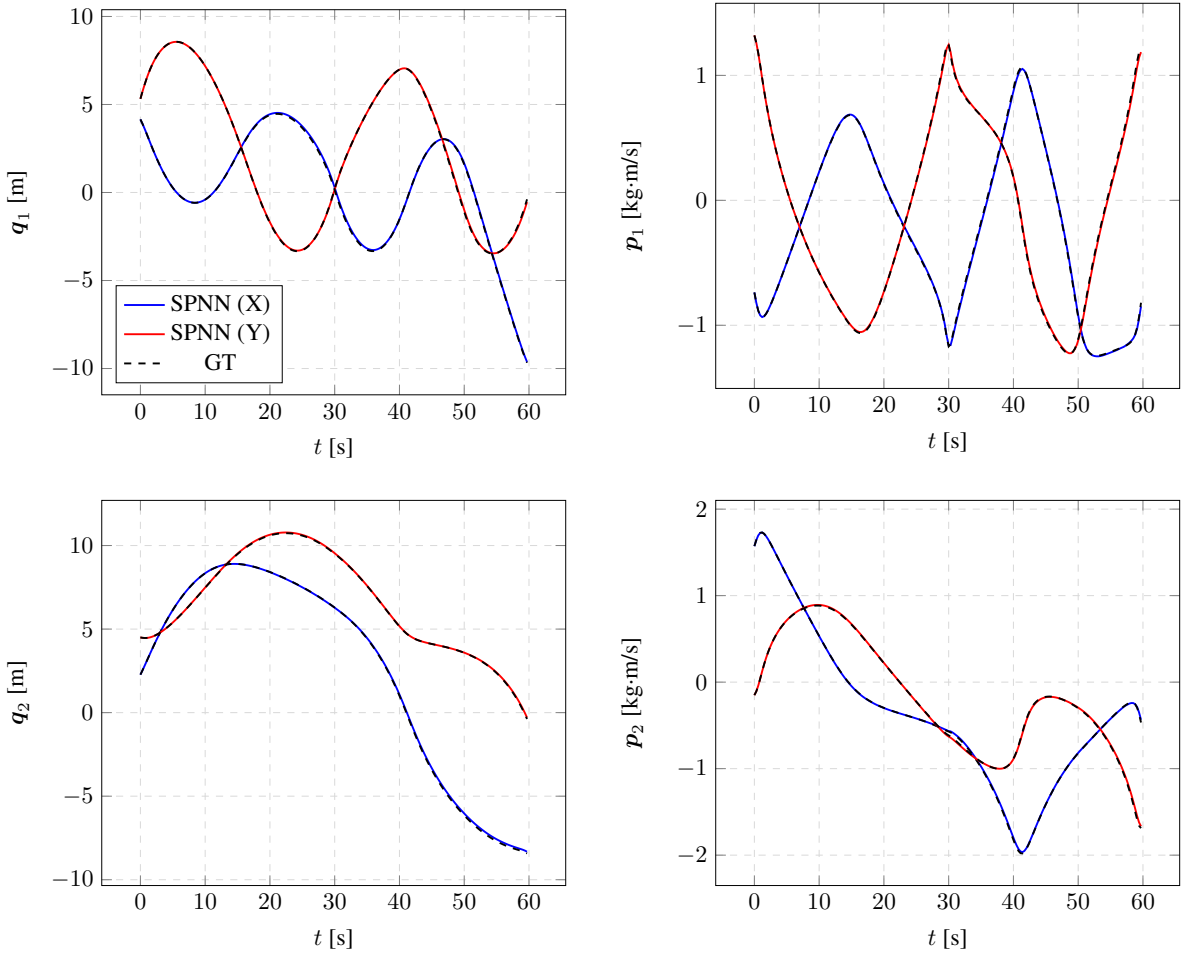

\myfigref{fig:double_results_time} shows the time evolution of the state variables (position, momentum and entropy) of each mass given by the solver and the neural net.

\mytabref{tab:double_results_error} shows the mean squared error of the data and degeneration loss terms for all the state variables of the double pendulum. The results are computed separately as the mean over all the train and test trajectories using \myeqref{eq:error_mean}.

\begin{table}
\caption{Mean squared error of the data loss ($\overline{\text{MSE}}^{\text{data}}$) and degeneracy loss ($\overline{\text{MSE}}^{\text{degen}}$) for all the state variables of the double pendulum.}
\label{tab:double_results_error}
\centering
\begin{tabular}{|l|l|l|l|l|l|}
\hline
\multicolumn{2}{|c|}{\multirow{2}{*}{State Variables}} & \multicolumn{2}{c|}{$\overline{\text{MSE}}^{\text{data}}$} & \multicolumn{2}{c|}{$\overline{\text{MSE}}^{\text{degen}}$}       \\ \cline{3-6} 
\multicolumn{2}{|c|}{}                                 & \multicolumn{1}{c|}{Train}   & \multicolumn{1}{c|}{Test}   & \multicolumn{1}{c|}{Train} & \multicolumn{1}{c|}{Test} \\ \hline
\multirow{2}{*}{$\bs{q}_1$ {[}m{]}}              & X        & $1.95\cdot 10^{-2}$          & $3.87\cdot 10^{-2}$         & $3.56\cdot 10^{-8}$        & $4.43\cdot 10^{-8}$       \\ \cline{2-6} 
                                            & Y        & $2.72\cdot 10^{-2}$          & $8.21\cdot 10^{-2}$         & $4.74\cdot 10^{-8}$        & $5.77\cdot 10^{-8}$       \\ \hline
\multirow{2}{*}{$\bs{q}_2$ {[}m{]}}              & X        & $2.04\cdot 10^{-2}$          & $3.65\cdot 10^{-2}$         & $9.28\cdot 10^{-8}$        & $8.55\cdot 10^{-8}$       \\ \cline{2-6} 
                                            & Y        & $2.72\cdot 10^{-2}$          & $3.73\cdot 10^{-2}$         & $3.55\cdot 10^{-8}$        & $5.01\cdot 10^{-8}$       \\ \hline
\multirow{2}{*}{$\bs{p}_1$ {[}kg·m/s{]}}         & X        & $6.43\cdot 10^{-4}$          & $1.33\cdot 10^{-4}$         & $4.00\cdot 10^{-8}$        & $7.08\cdot 10^{-8}$       \\ \cline{2-6} 
                                            & Y        & $1.06\cdot 10^{-3}$          & $4.06\cdot 10^{-3}$         & $1.21\cdot 10^{-7}$        & $1.40\cdot 10^{-7}$       \\ \hline
\multirow{2}{*}{$\bs{p}_2$ {[}kg·m/s{]}}         & X        & $4.88\cdot 10^{-4}$          & $9.84\cdot 10^{-4}$         & $6.00\cdot 10^{-8}$        & $4.58\cdot 10^{-8}$       \\ \cline{2-6} 
                                            & Y        & $8.47\cdot 10^{-4}$          & $1.79\cdot 10^{-4}$         & $9.76\cdot 10^{-8}$        & $1.20\cdot 10^{-7}$       \\ \hline
\multicolumn{2}{|l|}{$s_1$ {[}J/K{]}}                  & $1.21\cdot 10^{-5}$          & $3.51\cdot 10^{-5}$         & $1.31\cdot 10^{-7}$        & $2.06\cdot 10^{-7}$       \\ \hline
\multicolumn{2}{|l|}{$s_2$ {[}J/K{]}}                  & $1.22\cdot 10^{-5}$          & $3.18\cdot 10^{-5}$         & $2.40\cdot 10^{-7}$        & $2.95\cdot 10^{-7}$       \\ \hline
\end{tabular}
\end{table}

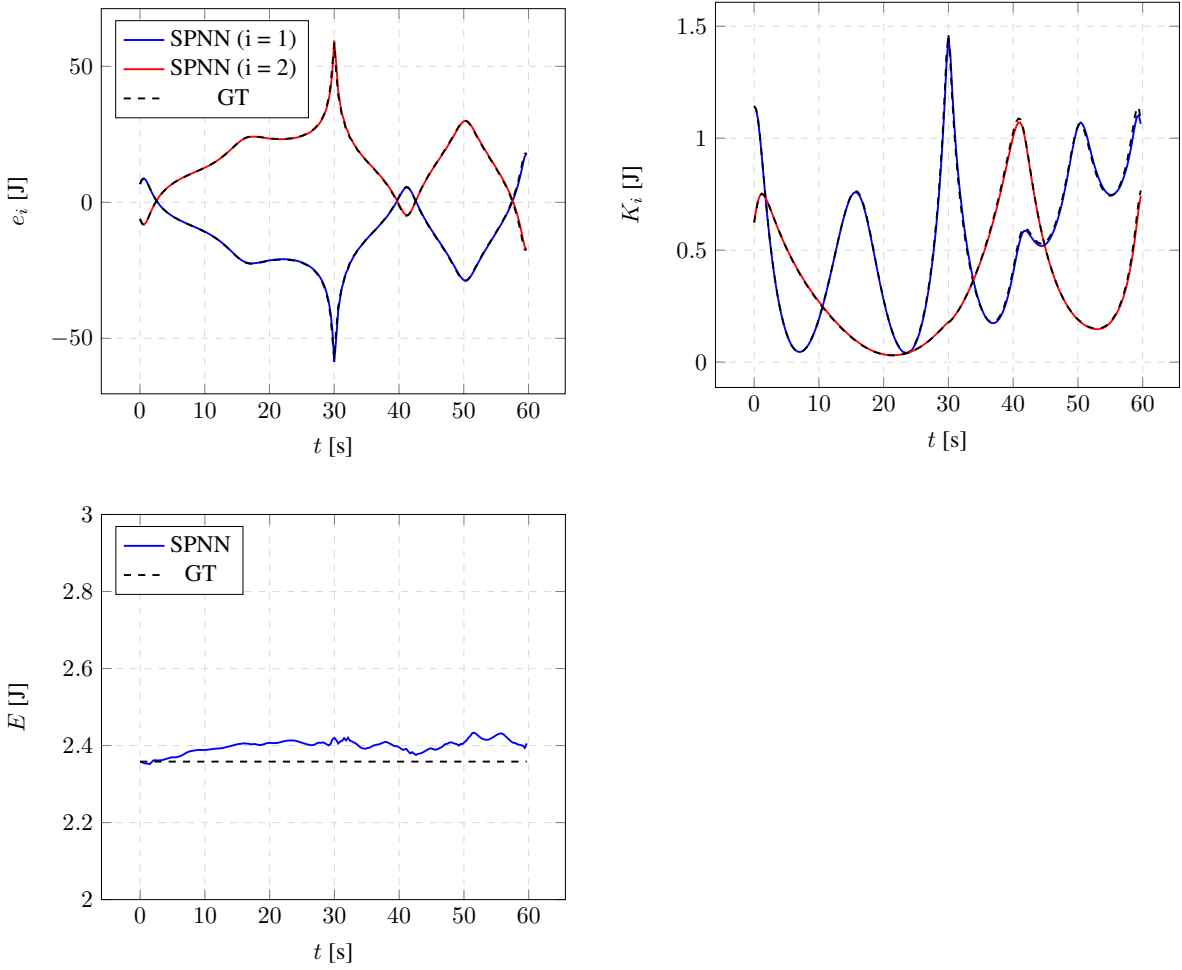
\begin{figure*}
\centering
\begin{tikzpicture}[scale=0.9]
%ei
\begin{axis}[name=plot1,
  grid=major, % Display a grid
  grid style={dashed,gray!30}, % Set the style
  xlabel={$t$ [s]},
  ylabel={$e_i$ [J]},
  legend pos=north west
]
  
  \foreach \F in {6}{  %{1,2,...,10}{
	\addplot [color=blue, thick] table [y=e1_net, x=tspan]{graphs/double_pendulum/double_energy_test_\F.txt};
	\ifthenelse{\F=6}{\addlegendentry{SPNN (i = 1)}}{}
	\addplot [color=red, thick] table [y=e2_net, x=tspan]{graphs/double_pendulum/double_energy_test_\F.txt};
	\ifthenelse{\F=6}{\addlegendentry{SPNN (i = 2)}}{}
	\addplot [color=black, thick, dashed, forget plot] table [y=e2_real, x=tspan]{graphs/double_pendulum/double_energy_test_\F.txt};
	\addplot [color=black, thick, dashed] table [y=e1_real, x=tspan]{graphs/double_pendulum/double_energy_test_\F.txt};
	\ifthenelse{\F=6}{\addlegendentry{GT}}{}
	}
\end{axis}

%E    at={($(plot1.east)+(2cm,0)$)},
\begin{axis}[name=plot3, at={($(plot1.below south east)+(0,-0.5cm)$)}, anchor=above north east,
  grid=major, % Display a grid
  grid style={dashed,gray!30}, % Set the style
  xlabel={$t$ [s]},
  ylabel={$E$ [J]},
  ymin=2.0,
  ymax=3.0,
  legend pos=north west
  ]
  
    \foreach \F in {6}{  %{1,2,...,10}{
	\addplot [color=blue, thick] table [y=e_net, x=tspan]{graphs/double_pendulum/double_energy_test_\F.txt};
	\ifthenelse{\F=6}{\addlegendentry{SPNN}}{}
	\addplot [color=black, thick, dashed] table [y=e_real, x=tspan]{graphs/double_pendulum/double_energy_test_\F.txt};
	\ifthenelse{\F=6}{\addlegendentry{GT}}{}
	}
\end{axis}

%ki
\begin{axis}[name=plot2, at={($(plot4.above north west)+(0,0.5cm)$)}, anchor=below south west,
  grid=major, % Display a grid 
  grid style={dashed,gray!30}, % Set the style
  xlabel={$t$ [s]},
  ylabel={$K_i$ [J]},
  ]
  
  \foreach \F in {6}{  %{1,2,...,10}{
	\addplot [color=blue, thick] table [y=k1_net, x=tspan]{graphs/double_pendulum/double_energy_test_\F.txt};
	%\ifthenelse{\F=6}{\addlegendentry{SPNN (i = 1)}}{}
	\addplot [color=black, thick, dashed] table [y=k1_real, x=tspan]{graphs/double_pendulum/double_energy_test_\F.txt};
	\addplot [color=red, thick] table [y=k2_net, x=tspan]{graphs/double_pendulum/double_energy_test_\F.txt};
	%\ifthenelse{\F=6}{\addlegendentry{SPNN (i = 2)}}{}
	\addplot [color=black, thick, dashed, forget plot] table [y=k2_real, x=tspan]{graphs/double_pendulum/double_energy_test_\F.txt};
	%\ifthenelse{\F=6}{\addlegendentry{GT}}{}
	}
\end{axis}

\end{tikzpicture}
\caption{Time evolution of the energy in a test trajectory of a double themo-elastic pendulum using a time-stepping solver (Ground Truth, GT) and the proposed GENERIC integration scheme (SPNN).}
\label{fig:double_energy_time}
\end{figure*}

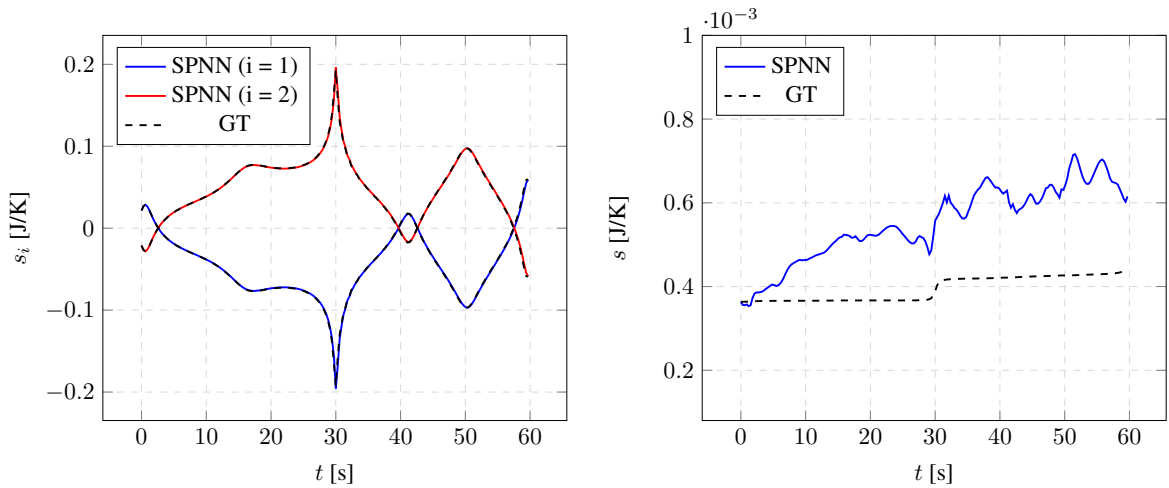
\begin{figure*}
\centering
\begin{tikzpicture}[scale=0.9]
%si
\begin{axis}[name=plot1,
  grid=major, % Display a grid
  grid style={dashed,gray!30}, % Set the style
  xlabel={$t$ [s]},
  ylabel={$s_i$ [J/K]},
  legend pos=north west
  ]
  
  \foreach \F in {6}{  %{1,2,...,10}{
	\addplot [color=blue, thick] table [y=s1_net, x=tspan]{graphs/double_pendulum/double_energy_test_\F.txt};
	\ifthenelse{\F=6}{\addlegendentry{SPNN (i = 1)}}{}
	\addplot [color=red, thick] table [y=s2_net, x=tspan]{graphs/double_pendulum/double_energy_test_\F.txt};
	\ifthenelse{\F=6}{\addlegendentry{SPNN (i = 2)}}{}
	\addplot [color=black, thick, dashed, forget plot] table [y=s2_real, x=tspan]{graphs/double_pendulum/double_energy_test_\F.txt};
	\addplot [color=black, thick, dashed] table [y=s1_real, x=tspan]{graphs/double_pendulum/double_energy_test_\F.txt};
	\ifthenelse{\F=6}{\addlegendentry{GT}}{}
	}
\end{axis}

%s
\begin{axis}[name = plot2, at={($(plot1.east)+(2cm,0)$)},anchor=west,
  grid=major, % Display a grid
  grid style={dashed,gray!30}, % Set the style
  xlabel={$t$ [s]},
  ylabel={$s$ [J/K]},
  ymin=8e-5,
  ymax=1e-3,
  legend pos=north west
  ]
  
  \foreach \F in {6}{  %{1,2,...,10}{
	\addplot [color=blue, thick] table [y=s_net, x=tspan]{graphs/double_pendulum/double_energy_test_\F.txt};
	\ifthenelse{\F=6}{\addlegendentry{SPNN}}{}
	\addplot [color=black, thick, dashed] table [y=s_real, x=tspan]{graphs/double_pendulum/double_energy_test_\F.txt};
	\ifthenelse{\F=6}{\addlegendentry{GT}}{}
	}
\end{axis}

\end{tikzpicture}
\caption{Time evolution of the entropy in a test trajectory of a double themo-elastic pendulum using a time-stepping solver (Ground Truth, GT) and the proposed GENERIC integration scheme (SPNN).}
\label{fig:double_entropy_time}
\end{figure*}

\myfigref{fig:double_energy_time} and \myfigref{fig:double_entropy_time} show the time evolution of the internal and kinetic energy (\myeqref{eq:double_energy}) and the entropy (\myeqref{eq:double_entropy}) respectively for the two pendulum masses ($i=1,2$). The total energy is conserved and the total entropy satisfies the entropy inequality, fulfilling the first and second laws of thermodynamics respectively. The mean error for both train and test trajectories is reported in \mytabref{tab:double_results_energy}.

\begin{table}[]
\caption{Mean squared error of the energy ($\overline{\text{MSE}}\;(E)$) and entropy ($\overline{\text{MSE}}\;(s)$) of the double pendulum.}
\centering
\label{tab:double_results_energy}
\begin{tabular}{|l|l|c|c|}
\hline
\multicolumn{2}{|c|}{Variable}      & Train               & Test                 \\ \hline
\multicolumn{2}{|l|}{$E$ {[}J{]}}   & $7.99\cdot 10^{-3}$ & $8.86\cdot 10^{-3}$  \\ \hline
\multicolumn{2}{|l|}{$S$ {[}J/K{]}} & $6.52\cdot 10^{-8}$ & $6.33\cdot 10^{-8}$ \\ \hline
\end{tabular}
\end{table}

\section{Couette flow of an Oldroyd-B fluid} \label{sec:visco}

\subsection{Description}

The second example is a shear (Couette) flow of an Oldroyd-B fluid model. This is a constitutive model for viscoelastic fluids, consisting of linear elastic dumbbells (representing polymer chains) immersed in a solvent.

The Oldroyd-B model arises in the modelling of flows of diluted polymeric solutions. This model can be obtained both from a purely macroscopic point of view as well as from a microscopic one, by modelling polymer chains as linear dumbbells diluted in a Newtonian substrate. Alternatively, it can also be obtained by considering the deviatoric part $\bs T$ of the stress tensor $\bs \sigma$ (the so-called extra-stress tensor), to be of the form
\begin{equation}
\boldsymbol{T}+\lambda_{1}\stackrel{\nabla}{\boldsymbol{T}}=\eta_{0}\left(\dot{\bs \gamma}+\lambda_{2}\stackrel{\nabla}{\dot{\bs \gamma}}\right),
\label{oldroydB:eq}
\end{equation}
where the triangle denotes the non-linear Oldroyd's upper-convected derivative \cite{Tim}. Coefficients $\eta_0$, $\lambda_1$ and $\lambda_2$ are model parameters. It is standard to denote the strain rate tensor by $\dot{\bs \gamma}=(\boldsymbol{\nabla}^s\bs  v)=\bs D$.

Finally, the stress in the solvent (denoted by a subscript $s$) and polymer (denoted by a subscript $p$) are given by 
\begin{equation*}
\boldsymbol{T}=\eta_{s}\dot{\bs \gamma}+\bs \tau,
\end{equation*}
so that %(see Appendix \ref{AppA})
\begin{equation*}
\bs \tau+\lambda_{1}\stackrel{\nabla}{\bs \tau}=\eta_{p}\dot{\bs \gamma},
\label{extrastress:eq}
\end{equation*}
which is the constitutive equation for the elastic stress.

\begin{figure}[h]
\centerline{
\input{./figures/visco/visco}
}
\caption{Couette flow in an Olroyd-B fluid.\label{fig:visco}}
\end{figure}
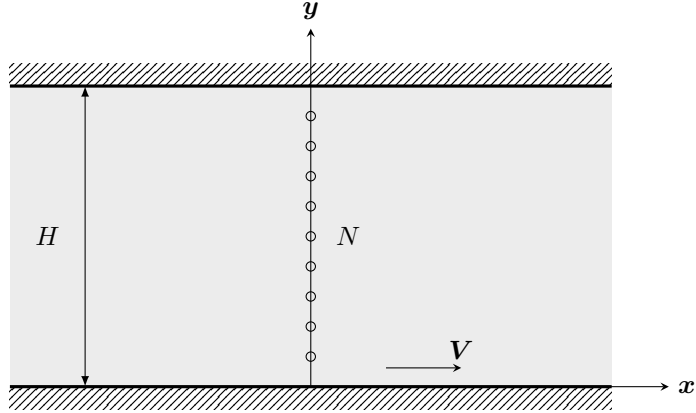

Pseudo-experimental data are obtained by the CONNFFESSIT technique \cite{laso1993calculation}, based on the Fokker-Plank equation \cite{le2009multiscale}. This equation is solved by converting it in its corresponding It\^o stochastic differential equation,
\begin{alignat}{1}\label{eq:visco_SDE}
 dr_x &=\left(\dpar{\bs v}{y}r_y-\frac{1}{2\text{We}}r_x\right)dt+\frac{1}{\sqrt{\text{We}}}dV_t, \nonumber\\
 dr_y &=-\frac{1}{2\text{We}}r_ydt+\frac{1}{\sqrt{\text{We}}}dW_t,
\end{alignat}
where $\bs v$ is the flow velocity, $\bs{r}=[r_x,\;r_y]^\top$, $r_x=r_x(y,t)$ the position vector and assuming a Couette flow so that $r_y=r_y(t)$ depends only on time, We stands for the Weissenberg number and $V_t$, $W_t$ are two independent one-dimensional Brownian motions. This equation is solved via Monte Carlo techniques, by replacing the mathematical expectation by the empirical mean.

The model relies on the microscopic description of the state of the dumbbells. Thus, it is particularly useful to base the microscopic description on the evolution of the conformation tensor $\bs c=\langle\bs{r}\bs{r}\rangle$, this is, the second moment of the dumbbell end-to-end distance distribution function. This tensor is in general not experimentally measurable and plays the role of an internal variable. The expected $xy$ stress component tensor will be given by
\begin{equation*}\label{eq:visco_tau}
\tau=\frac{\epsilon}{\text{We}}\frac{1}{K}\sum_{k=1}^Kr_xr_y,
\end{equation*}
where $K$ is the number of simulated dumbbells and $\epsilon=\frac{\nu_p}{\nu_p}$ is the ratio of the polymer to solvent viscosities.

The state variables selected for this problem are the position of the fluid on each node of the mesh, see Fig. \ref{fig:visco}, its velocity $\bs{v}$ in the x direction, internal energy $e$ and the conformation tensor shear component $\tau$,
\begin{equation*}\label{eq:visco_z}
\mathcal{S}=\{\bs{z}=(\bs{q},\bs{v},e,\tau)\in(\mathbb{R}^2\times\mathbb{R}\times\mathbb{R}\times\mathbb{R})\}.
\end{equation*}

The GENERIC matrices associated with each node of this physical system are the following
\begin{align}\label{eq:visco_LM}
    \mathsf{L} = \begin{bmatrix} 
           		0 & 0 & 1 & 0 & 0 & 0 \\
           		0 & 0 & 0 & 0 & 0 & 0 \\
           		-1 & 0 & 0 & 0 & 1 & -1 \\
           		0 & 0 & -1 & 0 & 0 & 0 \\
           		0 & 0 & 0 & 0 & 0 & 0 \\
              \end{bmatrix},
    \quad 
    \mathsf{M} = \begin{bmatrix} 
           		0 & 0 & 0 & 0 & 0 \\
           		0 & 0 & 0 & 0 & 0 \\
           		0 & 0 & 1 & 1 & 0 \\
           		0 & 0 & 1 & 1 & 0 \\
           		0 & 0 & 0 & 0 & 1 \\
              \end{bmatrix}.   
\end{align}

In order to simulate a measurement of real captured data, Gaussian noise is added to the state vector, computed as a random variable following a normal distribution with zero mean and standard deviation proportional to the standard deviation of the database $\sigma_{\bs{z}}$ and noise level $\nu$,
\begin{equation}
\bs{z}^{\text{GT}}_{noise}=\bs{z}^{\text{GT}}+\nu\cdot\sigma_{\bs{z}}\cdot\mathcal{N}(0,1)
\end{equation}

The results of both the noise-free and the noisy database are compared with two different network architectures:

\begin{itemize}
\item Unconstrained network: This architecture is the same as the proposed network but removing the degeneracy conditions of the energy and entropy, \myeqref{eq:cost_degen}, in the loss function. These conditions ensure the thermodynamic consistency of the resulting integrator, so not including them affects negatively in the accuracy of the results, as will be seen.
\item Black-Box network: In this case, no GENERIC architecture is imposed, acting as a black-box integrator trained to directly predict the state vector time evolution $\bs{z}_{t+1}$ from the previous time step $\bs{z}_t$. This naive approach is shown to be inappropriate, as no physical restrictions are given to the model.
\end{itemize}

\subsection{Database and Hyperparameters}

The training database for this Olroyd-B model is generated in MATLAB with a multiscale approach \cite{le2009multiscale} in the dimensionless form. The fluid is discretized in the vertical direction with $N=100$ elements (101 nodes) in a total height of $H=1$. A total of 10,000 dumbells were considered at each nodal location in the model. The lid velocity is set to $V=1$, the viscolastic Weissenberg number We $=1$ and Reynolds number of Re $ = 0.1$. The simulation time of the movement is $T = 1$ in time increments of $\Delta t = 0.0067$ ($N_T=150$ snapshots). 

The database consisted of the state vector (\myeqref{eq:visco_z}) of the 100 nodes trajectories (excluding the node at $h=H$, for which a no-slip condition $v=0$ has been imposed). This database is split in 80 train trajectories and 20 test trajectories. 

The net input and output size is $N_{\text{in}} = 5$ and $N_{\text{out}} = 2N_{\text{in}}^2 = 50$. The number of hidden layers is $N_h = 5$ with ReLU activation functions and linear in the last layer. It is initialized according to the Kaiming method \cite{he2015delving}, with normal distribution and the optimizer used is Adam \cite{kingma2014adam}, with a weight decay of $\lambda_r=10^{-5}$ and data loss weight of $\lambda_d = 10^3$. A multistep learning rate scheduler is used, starting in $\eta=10^{-3}$ and decaying by a factor of $\gamma=0.1$ in epochs 500 and 1000. The training process ends when a fixed number of epochs $n_{epoch}=1500$ is reached. The same parameters are considered also for the noisy database network ($\nu=1\%$) and the unconstrained network.

The black-box network training parameters are analogous to the structure-preserving network, except for the output size $N_{\text{out}} = N_{\text{in}} = 5$. Several network architectures were tested, and the lowest error is achieved with  $N_h = 5$ hidden layers and 25 neurons each layer.

\begin{figure*}[h!]
\centering
\begin{tikzpicture}[scale=0.9]
\begin{semilogyaxis}[name = plot1,
  grid=major, % Display a grid
  grid style={dashed,gray!30}, % Set the style
  xlabel={$\text{Epoch}$ [-]},
  ylabel={$\text{Loss}$ [-]},
  ]
  
	\addplot [red,thick] table [y=loss_error, x=epoch]{graphs/visco/viscolastic_log.txt};
	\addlegendentry{$\mathcal{L}^\text{data}$}
	\addplot [blue,thick] table [y=loss_degeneracy, x=epoch]{graphs/visco/viscolastic_log.txt};
	\addlegendentry{$\mathcal{L}^\text{degen}$}
\end{semilogyaxis}

\end{tikzpicture}
\caption{Loss evolution of data and degeneracy constraints for each epoch of the neural network training process of the Couette flow example.}
\label{fig:visco_log}
\end{figure*}
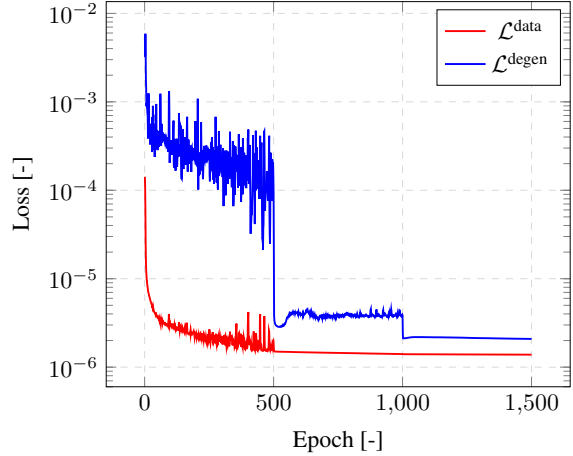

The time evolution of the data $\mathcal{L}^\text{data}$ and degeneracy $\mathcal{L}^\text{degen}$ loss terms for each training epoch are shown in \myfigref{fig:visco_log}.

\subsection{Results}

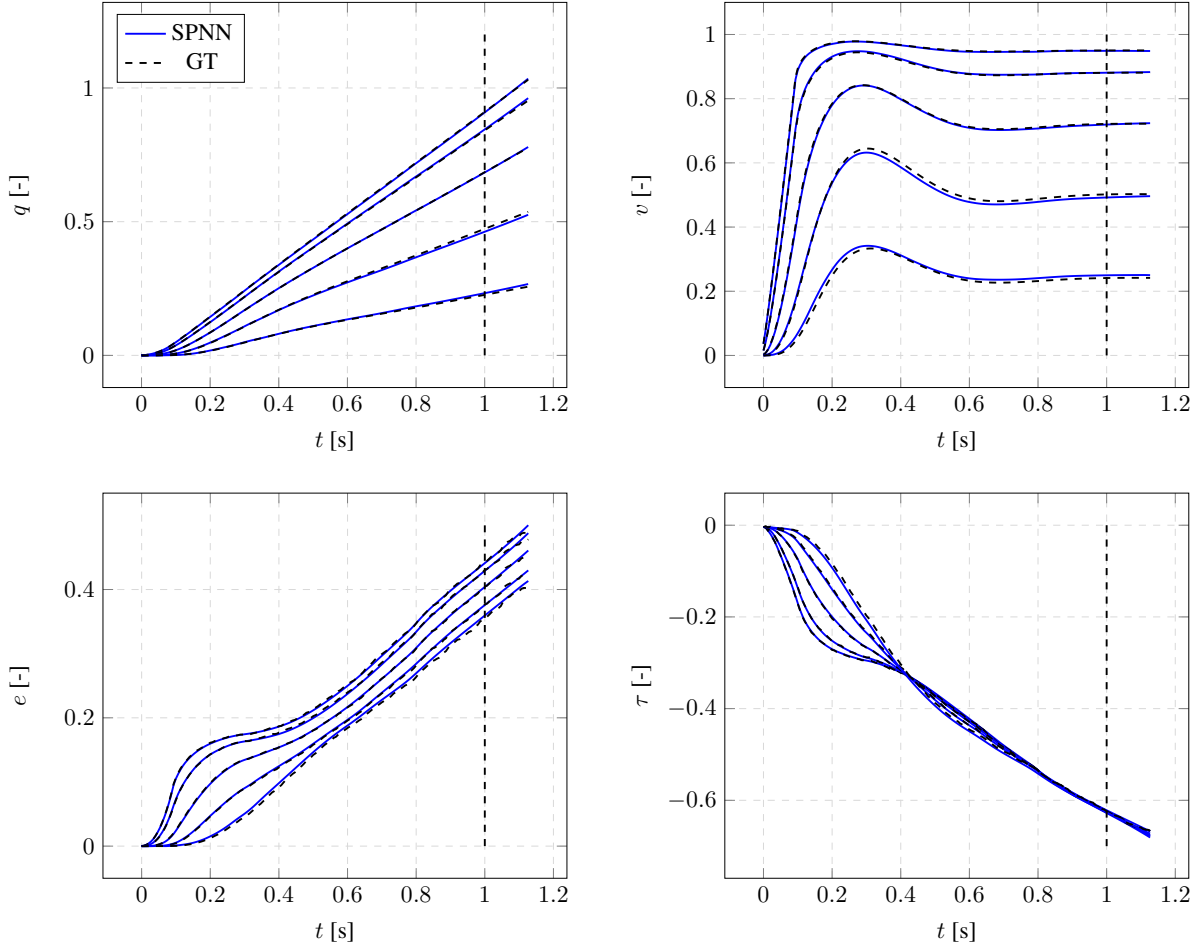
\begin{figure*}[h]
\centering
\begin{tikzpicture}[scale=0.9]
% qx
\begin{axis}[name=plot1,
  grid=major, % Display a grid
  grid style={dashed,gray!30}, % Set the style
  xlabel={$t$ [s]},
  ylabel={$q$ [-]},
  legend pos=north west
  ]
  
  \foreach \F in {15,19,2,6,10}{  %{1,2,...,20}{
	\addplot [color=blue, thick] table [y=qx_net, x=tspan]{graphs/visco/viscolastic_results_test_\F.txt};
	\ifthenelse{\F=15}{\addlegendentry{SPNN}}{}
	\addplot [color=black, thick, dashed] table [y=qx_real, x=tspan]{graphs/visco/viscolastic_results_test_\F.txt};
	\ifthenelse{\F=15}{\addlegendentry{GT}}{}
	}
	\addplot +[color=black,mark=none,thick,dashed] coordinates {(1, 0) (1, 1.2)};
\end{axis}

%e    at={($(plot1.east)+(2cm,0)$)},
\begin{axis}[name=plot3, at={($(plot1.below south east)+(0,-0.5cm)$)}, anchor=above north east,
  grid=major, % Display a grid
  grid style={dashed,gray!30}, % Set the style
  xlabel={$t$ [s]},
  ylabel={$e$ [-]},
  ]
  
  \foreach \F in {15,19,2,6,10}{  %{1,2,...,20}{
	\addplot [color=blue, thick] table [y=e_net, x=tspan]{graphs/visco/viscolastic_results_test_\F.txt};
	%\ifthenelse{\F=15}{\addlegendentry{SPNN}}{}
	\addplot [color=black, thick, dashed] table [y=e_real, x=tspan]{graphs/visco/viscolastic_results_test_\F.txt};
	%\ifthenelse{\F=15}{\addlegendentry{GT}}{}
	}
	\addplot +[color=black,mark=none,thick,dashed] coordinates {(1, 0) (1, 0.5)};
\end{axis}

%tau
\begin{axis}[name=plot4, at={($(plot3.right of north east)+(0.7cm,0)$)}, anchor=left of north west,
  grid=major, % Display a grid
  grid style={dashed,gray!30}, % Set the style
  xlabel={$t$ [s]},
  ylabel={$\tau$ [-]},
  ]
  
  \foreach \F in {15,19,2,6,10}{  %{1,2,...,20}{
	\addplot [color=blue, thick] table [y=tau_net, x=tspan]{graphs/visco/viscolastic_results_test_\F.txt};
	%\ifthenelse{\F=15}{\addlegendentry{SPNN}}{}
	\addplot [color=black, thick, dashed] table [y=tau_real, x=tspan]{graphs/visco/viscolastic_results_test_\F.txt};
	%\ifthenelse{\F=15}{\addlegendentry{GT}}{}
	}
	\addplot +[color=black,mark=none,thick,dashed] coordinates {(1, 0) (1, -0.7)};
\end{axis}

%vx
\begin{axis}[name=plot2, at={($(plot4.above north west)+(0,0.5cm)$)}, anchor=below south west,
  grid=major, % Display a grid 
  grid style={dashed,gray!30}, % Set the style
  xlabel={$t$ [s]},
  ylabel={$v$ [-]},
  %legend style={at={(0.3,1.35)},anchor=west},
  %legend pos=outer north east
  ]
  
  \foreach \F in {15,19,2,6,10}{  %{1,2,...,20}{
	\addplot [color=blue, thick] table [y=vx_net, x=tspan]{graphs/visco/viscolastic_results_test_\F.txt};
	%\ifthenelse{\F=15}{\addlegendentry{SPNN}}{}
	\addplot [color=black, thick, dashed] table [y=vx_real, x=tspan]{graphs/visco/viscolastic_results_test_\F.txt};
	%\ifthenelse{\F=15}{\addlegendentry{GT}}{}
	}
	\addplot +[color=black,mark=none,thick,dashed] coordinates {(1, 0) (1, 1)};
\end{axis}

\end{tikzpicture}
\caption{Time evolution of the state variables in five test nodes of a Couette flow using a solver (Ground Truth, GT) and the proposed GENERIC integration scheme (Net). The dotted vertical line represent the simulation time $T=1$ of the training dataset.}
\label{fig:visco_results_time}
\end{figure*}

\myfigref{fig:visco_results_time} shows the time evolution of the state variables (position $q$, velocity $v$, internal energy $e$ and conformation tensor shear component $\tau$) given by the solver and the neural net. There is a good agreement between both plots. Moreover, the proposed scheme is able to predict the time evolution of the flow for several snapshots beyond the training simulation time $T=1$, as shown in the same figure.

\mytabref{tab:visco_results_error} show the mean squared error of the data and degeneration loss terms for all the state variables of the Couette flow of an Olroyd-B fluid. The results are computed separately as the mean over all the train and test trajectories using \myeqref{eq:error_mean}.

\begin{table}
\caption{Mean squared error of the data loss ($\overline{\text{MSE}}^{\text{data}}$) and degeneracy loss ($\overline{\text{MSE}}^{\text{degen}}$) for all the state variables of the Couette flow.}
\label{tab:visco_results_error}
\centering
\begin{tabular}{|l|l|c|c|c|c|}
\hline
\multicolumn{2}{|c|}{\multirow{2}{*}{State Variables}} & \multicolumn{2}{c|}{$\overline{\text{MSE}}^{\text{data}}$} & \multicolumn{2}{c|}{$\overline{\text{MSE}}^{\text{degen}}$} \\ \cline{3-6} 
\multicolumn{2}{|c|}{}                                 & Train                        & Test                        & Train                   & Test                   \\ \hline
\multirow{2}{*}{$q$ {[}-{]}}            & X            & $5.40\cdot 10^{-6}$          & $6.29\cdot 10^{-6}$         & $1.72\cdot 10^{-7}$     & $1.96\cdot 10^{-7}$    \\ \cline{2-6} 
                                        & Y            & $0.00$                       & $0.00$                      & $0.00$                  & $0.00$                 \\ \hline
\multicolumn{2}{|l|}{$v$ {[}-{]}}                      & $3.23\cdot 10^{-5}$          & $4.75\cdot 10^{-5}$         & $1.19\cdot 10^{-6}$     & $1.48\cdot 10^{-6}$    \\ \hline
\multicolumn{2}{|l|}{$e$ {[}-{]}}                      & $7.85\cdot 10^{-6}$          & $6.60\cdot 10^{-6}$         & $7.06\cdot 10^{-7}$     & $9.11\cdot 10^{-7}$    \\ \hline
\multicolumn{2}{|l|}{$\tau$ {[}-{]}}                   & $2.36\cdot 10^{-5}$          & $1.26\cdot 10^{-5}$         & $1.07\cdot 10^{-6}$     & $1.31\cdot 10^{-6}$    \\ \hline
\end{tabular}
\end{table}

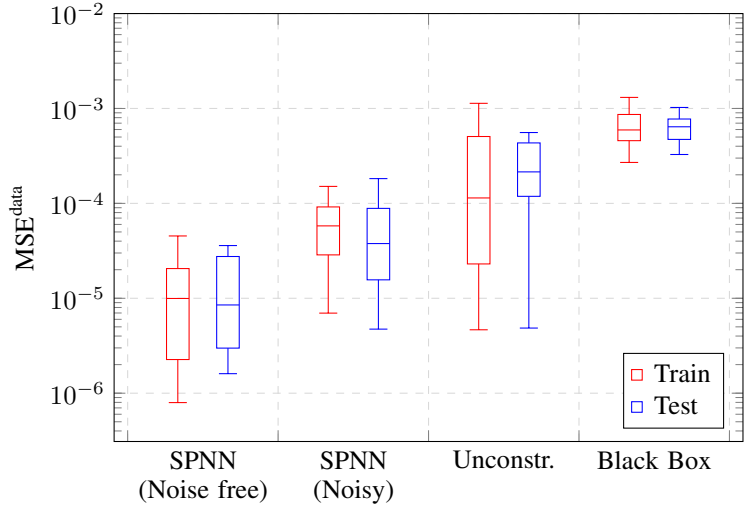
\begin{figure*}[h!]
\centering
\begin{tikzpicture}
%\pgfplotsset{width=\textwidth, height=8cm}
  \begin{semilogyaxis}[
  boxplot/draw direction=y,
  grid=major, % Display a grid
  grid style={dashed,gray!30}, % Set the style
  ylabel={$\text{MSE}^{\text{data}}$},
  cycle list={{red},{blue}},
  boxplot={draw position={1/3 + floor(\plotnumofactualtype/2) + 1/3*mod(\plotnumofactualtype,2)},
  	box extend=0.15,},
  x=2cm,
  xtick={0,1,2,...,50},
  x tick label as interval,
  xticklabels={{SPNN\\(Noise free)},{SPNN\\(Noisy)},{Unconstr.},{Black Box}},
  x tick label style={text width=2.5cm,align=center},
  custom legend,
  legend pos= south east,
  legend cell align=left,
  ymax=0.01
  ]
  
  	% Noiseless
  	% Train
    \addplot+ [boxplot] table [y=mse_train] {graphs/visco/viscolastic_error_noiseless_train.txt};
    \addlegendentry{~Train}
    % Test
3    \addplot+ [boxplot] table [y=mse_test] {graphs/visco/viscolastic_error_noiseless_test.txt};
    \addlegendentry{~Test}
    
    % Noise
  	% Train
    \addplot+ [boxplot] table [y=mse_train] {graphs/visco/viscolastic_error_noise_train.txt};
    % Test
    \addplot+ [boxplot] table [y=mse_test] {graphs/visco/viscolastic_error_noise_test.txt};
        
    % No Deg
  	% Train
    \addplot+ [boxplot] table [y=mse_train] {graphs/visco/viscolastic_error_noDegen_train.txt};
    % Test
    \addplot+ [boxplot] table [y=mse_test] {graphs/visco/viscolastic_error_noDegen_test.txt};
    
    % Black Box
  	% Train
    \addplot+ [boxplot] table [y=mse_train] {graphs/visco/viscolastic_error_BlackBox_train.txt};
    % Test
    \addplot+ [boxplot] table [y=mse_test] {graphs/visco/viscolastic_error_BlackBox_test.txt};
    
  \end{semilogyaxis}
\end{tikzpicture}
\caption{Box plots for the data integration mean squared error ($\text{MSE}^{\text{data}}$) of the Couette flow in both train and test cases.}
\label{fig:visco_box}
\end{figure*}

\myfigref{fig:visco_box} shows a box plot of the data error ($\text{MSE}^{\text{data}}$) for the train and test sets in the four studied architectures. The results of the structure-preserving neural network outperform the other two approaches even with noisy training data. The error of the unconstrained neural network is greater than one order of magnitude than our approach, proving the importance of the degeneracy conditions in the GENERIC formulation. Last, the naive black-box approach shows the worst performance of the four networks, as no physical restriction is considered.

With respect to our previous work\cite{gonzalez2019thermodynamically}, that employed a piece-wise linear regression approach, these examples show similar levels of accuracy, but a much greater level of robustness. For instance, this same example was included in the mentioned reference. However, in that case, the problem had to be solved with the help of a reduced order model with only six degrees of freedom, due to the computational burden of the approach. In our former approach, the GENERIC structure was identified by piece-wise linear regression for each of the few global {\it modes} of the approximation. So to speak, in that case, we {\it learnt} the characteristics of the flow. Here, on the contrary, the net is able to find an approximation for any velocity value at the 101 nodes of the mesh---say, fluid particles---without any difficulty. In this case, we are {\it learning} the behavior of fluid particles.  It will be interesting, however, to study to what extent the employ of variational autoencoders, as in Bertalan et al.\cite{KevrekidisHamiltonian}, could help in solving more intricate models. Autoencoders help in determining the actual number of degrees of freedom needed to represent a given physical phenomenon. 

\section{Conclusions}\label{sec:conc}

In this work we have presented a new methodology to ensure thermodynamic consistency in the deep learning of physical phenomena. In contrast to existing methods, this methodology does not need to know in advance any information related to balance equations or the precise form of the PDE governing the phenomena at hand. The method is constructed on top of the right thermodynamic principles that ensure the fulfillment of the energy dissipation and entropy production. It is valid, therefore, for conservative as well as dissipative systems, thus overcoming previous approaches in the field.

When compared with our previous works in the field (see Gonzalez et al. \cite{gonzalez2019thermodynamically}), the present methodology showed to be more robust, allowing us to find approximations for systems with orders of magnitude more degrees of freedom. This new approach is also less computationally demanding. For the double pendulum case, the snapshot optimization of the GENERIC matrices proposed in \cite{gonzalez2019thermodynamically} has a measured performance of 10 min per trajectory, which add up to 400 minutes considering the 40 studied trajectories, whereas our new neural-network approach trains in only 73.18 minutes. The computational time of the other examples is shown in \mytabref{tab:time}

\begin{table}[]
\centering
\caption{Computation training time of the proposed algorithm for the two reported examples in the noise free networks.}
\label{tab:time}
\begin{tabular}{|l|c|c|}
\hline
\multicolumn{1}{|c|}{Example} & Epoch Time     & Total Time \\ \hline
Double Pendulum               & 2.44 s/epoch   & 73.18 min  \\ \hline
Couette Flow                  & 1.22 s/epoch   & 30.53 min  \\ \hline
\end{tabular}
\end{table}

The reported results show good agreement between the network output and the synthetic ground truth solution, even with moderate noisy data. We have also shown the importance of including the degeneracy conditions of the GENERIC formulation to the neural network constraints, as it ensures the thermodynamical consistency of the integrator. The structure-preserving neural network outperforms other naive black-box approaches, since the physical constraints act as an inductive bias, facilitating the learning process. However, the error can be reduced using several techniques: 
\begin{itemize}
\item \textbf{Database}: As a general method of increasing the precision of an Euler integration scheme, the time step $\Delta t$ can be decreased so the total number of snapshots is increased. On the contrary, the database will be larger, slowing the training process. The same way, the database can be enriched with a wider variety of cases, improving the net predictive capabilities.

\item \textbf{Integration Scheme}: A higher order Runge-Kutta integration scheme could be introduced in \myeqref{eq:generic_disc} in order to get higher solution accuracy\cite{wang1998runge}. However, it requires several forward passes through the neural net for each time step, incrementing the complexity of the integration scheme and the training process. Additionally, GENERIC-based integration schemes have showed very good performance even for first-order approaches.\cite{romero2009thermodynamically}

\item \textbf{Net Architecture}: To increase the computational power of the net, more and larger hidden layers $N_h$ can be added. However, this could lead to a more over-fitted solution which limit the prediction power and versatility of the net. It also increases the computational cost of both the training process and the testing of the net.

\item \textbf{Training Hyperparameters}: The neural networks trained in this work could be optimized using several hyperparameter tuning methods such as random search, Bayesian optimization or gradient-based optimization to get a more efficient solution.
\end{itemize}

Several open questions remain as a future work. A more exhaustive analysis can be performed to evaluate the influence of noisy data to the integrator evolution, in order to add robustness to the method and even predict wider simulation times using incremental learning \cite{li2017learning, choi2019autoencoder}.

\section*{Acknowledgements}

This project has been partially funded by the ESI Group through the ESI Chair at ENSAM Arts et Metiers Institute of Technology, and through the project 2019-0060 ``Simulated Reality'' at the University of Zaragoza. The support of the Spanish Ministry of Economy and Competitiveness through grant number CICYT-DPI2017-85139-C2-1-R and by the Regional Government of Aragon and the European Social Fund, are also gratefully acknowledged.

% Bibliography

%\section*{Bibliography}
%
%\bibliography{bibNEM}
%\bibliographystyle{unsrt}

\end{document}

%% file: figures/neural_net/neural_net.tex
\begin{tikzpicture}

	\tikzstyle{inputNode}=[draw,circle,minimum size=10pt,inner sep=0pt]
	\tikzstyle{stateTransition}=[-stealth, thick]	
	
	% Neuron
	\node[draw,circle,minimum size=25pt,inner sep=0pt] (node_sum) at (2-1,0) {$\Sigma$};
	\node[draw,circle,minimum size=25pt,inner sep=0pt] (node_act) at (3.25-1,0) {$\sigma_j$};

	\node[inputNode,minimum size=25pt] (in_b) at (-1, 1.5) {$+1$};
	\node[inputNode,minimum size=25pt] (in_1) at (-1, 0) {$x_1^{[l-1]}$};
	\node[inputNode,minimum size=25pt] (in_n) at (-1, -1.5) {$x_i^{[l-1]}$};

	\draw[stateTransition] (in_b) to[out=0,in=120] node [midway,above] {$b_j^{[l]}$} (node_sum);
	\draw[stateTransition] (in_1) -- (node_sum) node [midway,above] {$w_{1,j}^{[l]}$};
	\draw[stateTransition] (in_n) to[out=0,in=240] node [midway,below] {$w_{i,j}^{[l]}$} (node_sum);
	\draw[stateTransition] (node_sum) -- (node_act) node [midway,above] {};
	\draw[stateTransition] (node_act) -- (4.5-1,0) node [midway,above] {$x^{[l]}_j$};
	\node (dots) at (0-1, -0.75) {$\vdots$};	
	\node (dots) at (1-1, -0.6) {$\vdots$};	
	
	% Net
	\node[inputNode, thick] (i1) at (6, 0.75) {};
	\node[inputNode, thick] (i3) at (6, -0.75) {};
	
	\node[inputNode, thick] (h1) at (8, 1.5) {};
	\node[inputNode, thick] (h2) at (8, 0.75) {};
	\draw[dashed] (8-0.3, 0.75-0.3) -- (8-0.3, 0.75+0.3) -- (8+0.3, 0.75+0.3) -- (8+0.3, 0.75-0.3) -- (8-0.3, 0.75-0.3);
	%\node[inputNode, thick] (h3) at (8, 0) {};
	\node (dots) at (8, 0) {$\vdots$};
	\node[inputNode, thick] (h4) at (8, -0.75) {};
	\node[inputNode, thick] (h5) at (8, -1.5) {};
	
	\node[inputNode, thick] (o1) at (10, 0.75) {};
	\node[inputNode, thick] (o2) at (10, -0.75) {};
	
	\draw[stateTransition] (5, 0.75) -- node[above] {$I_1$} (i1);
	\node (dots) at (5.4, 0.2) {$\vdots$};
	\draw[stateTransition] (5, -0.75) -- node[above] {$N_{in}$} (i3);
	
	\draw[stateTransition] (i1) -- (h1);
	\draw[stateTransition] (i1) -- (h2);
	%\draw[stateTransition] (i1) -- (h3);
	\draw[stateTransition] (i1) -- (h4);
	\draw[stateTransition] (i1) -- (h5);
	\draw[stateTransition] (i3) -- (h1);
	\draw[stateTransition] (i3) -- (h2);
	%\draw[stateTransition] (i3) -- (h3);
	\draw[stateTransition] (i3) -- (h4);
	\draw[stateTransition] (i3) -- (h5);
	
	\draw[stateTransition] (h1) -- (o1);
	\draw[stateTransition] (h1) -- (o2);
	\draw[stateTransition] (h2) -- (o1);
	\draw[stateTransition] (h2) -- (o2);
	%\draw[stateTransition] (h3) -- (o1);
	%\draw[stateTransition] (h3) -- (o2);
	\draw[stateTransition] (h4) -- (o1);
	\draw[stateTransition] (h4) -- (o2);
	\draw[stateTransition] (h5) -- (o1);
	\draw[stateTransition] (h5) -- (o2);
	
	\node[above=of i1, align=center] (l1) {Input \\ layer};
	\node[right=2.3em of l1, align=center] (l2) {Hidden \\ layers ($N_h$)};
	\node[right=2.3em of l2, align=center] (l3) {Output \\ layer};
	\node[left=8.5em of l1, align=center] (l4) {Layer $l$\\ Neuron $j$};
	
	\draw[stateTransition] (o1) -- node[above] {$O_1$} (11, 0.75);
	\node (dots) at (10.6, 0.2) {$\vdots$};
	\draw[stateTransition] (o2) -- node[above] {$N_{out}$} (11, -0.75);
	
\end{tikzpicture}

%% file: figures/algorithm_blocks/algorithm_blocks.tex
%\documentclass[tikz,14pt,border=10pt]{standalone}
%\usetikzlibrary{shapes,arrows}
%\usetikzlibrary{calc}
%\begin{document}

% Defining string as labels of certain blocks.
%\newcommand{\dpar}[2]{\frac{\partial #1}{\partial #2}}
%\newcommand{\bs}[1]{\boldsymbol{#1}}
\newcommand{\suma}{\Large$+$}
\begin{tikzpicture}[auto, thick, >=triangle 45]
\draw
	% Define blocks
	node at (0,0) (input) {} 
	node [draw, rectangle, minimum height = 3em, minimum width = 4em] at (2.5,0) (NN) {NN}
	node [draw, rectangle, minimum height = 3em, minimum width = 7em] at (7,0) (GENERIC) {GENERIC}
    node [draw, rectangle, minimum height = 3em, minimum width = 3em] at (10.5,0) (SSE) {SSE}
    node [draw, circle, node distance = 2cm] at (13,0) (sum) {\suma}
    node [draw, rectangle, minimum height = 3em, minimum width = 3em] at (15,0) (sumL) {Loss};
    % Joining blocks. 
	\draw[->](input) -- node {$\bs{z}_n$}(NN);
	\draw[->](NN) -- node {$\bs{A}^{\text{net}},\bs{B}^{\text{net}}$} (GENERIC);
	\draw[->](GENERIC) -- node {$\bs{z}_{n+1}^{\text{net}}$} (SSE);
	\draw[->](SSE) -- node {$\mathcal{L}^{\text{data}}$} (sum);
	\draw[->](sum) -- (sumL);	
	
	\node at ($(NN)+(0,-1.8)$)(wb) {$\bs{\phi}=\bs{w},\bs{b}$};
	\draw[->](wb) -- (NN);
	
	\node at ($(GENERIC)+(0,-1.8)$)(dt) {$\Delta t,L,M$};
	\draw[->](dt) -- (GENERIC);
	
	\node at ($(SSE)+(0,-1.8)$)(GT) {$z_{n+1}^{\text{GT}}$};
	\draw[->](GT) -- (SSE);		
	
	\node at ($(sum)+(0,-1.8)$)(Ldeg) {$\mathcal{L}^{\text{reg}}$};
	\draw[->](Ldeg) -- (sum);
	
	\draw[->] (sumL.south) -- ++(0,-2.5cm) -| node [near start] {Net Update: $\bs{\phi}\leftarrow\bs{\phi}-\eta\dpar{\mathcal{L}}{\bs{\phi}}$} ($(wb.south)$);
	
	\draw[->] (GENERIC.north) -- ++(0,1cm) -| node [near start] {$\mathcal{L}^{\text{degen}}$} ($(sum.north)$);	

\end{tikzpicture}
%\end{document}

%% file: figures/double_pendulum/double_pendulum.tex
\begin{tikzpicture}

% Origen
\node[draw,circle,scale=0.25*1.5] (origin) at (0,0) {};
\node [fill,pattern=north east lines,draw=none,minimum width=0.8cm*1.5,minimum height=0.2cm*1.5,anchor=south] (ground) at ($(origin) + (0,0.15*1.5)$)  {};
\draw[very thick] (-0.4*1.5,0.15*1.5) -- (0.4*1.5,0.15*1.5);
\node (ground_left) at ($(ground) + (0.3,-0.02)$) {};
\node (ground_right) at ($(ground) + (-0.3,-0.02)$) {};
\draw (ground_left) -- (origin);
\draw (ground_right) -- (origin);

% Coords
\node (x) at (1,0) {$\bs{x}$};
\node (y) at (0,-1) {$\bs{y}$};
\draw[-stealth] (origin)  -- (x);
\draw[-stealth] (origin)  -- (y);

% Bolas 1 y 2
\node[draw,thick,circle,scale=1.5] (ball_1) at (1*1.5,-1.5*1.5) {};
\node[draw=none,below=.25cm*1.5] at (ball_1) {$m_1$};
\node[draw=none] (spring_1) at (0.1,-1*1.5) {$\lambda_1, C_1$};
\node[draw,thick,circle,scale=1.5] (ball_2) at (3.5*1.5,-3*1.5) {};
\node[draw=none,above=.25cm*1.5] at (ball_2) {$m_2$};
\node[draw=none] (spring_2) at (2.5*1.5,-1.9*1.5) {$\lambda_2, C_2$};

% Velocidades
\node (p_1) at ($(ball_1) + (0.5*1.5,1*1.5)$) {$\bs{p}_1$};	
\node (p_2) at ($(ball_2) + (-1.5*1.5,0.3*1.5)$) {$\bs{p}_2$};	
\draw[-stealth] (ball_1) -- (p_1);
\draw[-stealth] (ball_2) -- (p_2);

% Muelles
\draw[decorate,decoration={aspect=0.5, segment length=1mm, amplitude=1mm,coil}] (origin) -- (ball_1);
\draw[decorate,decoration={aspect=0.5, segment length=1.5mm, amplitude=1mm,coil}] (ball_1) -- (ball_2);

\end{tikzpicture}

%% file: figures/visco/visco.tex
\begin{tikzpicture}

% Bottom
\node (bot) at (0,0) {};
\node (bot_left) at ($(bot) + (-4,-0.3)$) {};
\node (bot_right) at ($(bot) + (4,0)$) {};
\draw [pattern=north east lines,draw=none] (bot_left) rectangle (bot_right);

% Top
\node (top) at (0,4) {};
\node (top_left) at ($(top) + (-4,+0.3)$) {};
\node (top_right) at ($(top) + (4,0)$) {};
\draw [pattern=north east lines,draw=none] (top_left) rectangle (top_right);

% Fluid
\node (fluid_left) at ($(bot) + (-4,0)$) {};
\node (fluid_right) at ($(top) + (4,0)$) {};
\filldraw [draw=none, color=lightgray!30] (fluid_left) rectangle (fluid_right);

\draw[very thick] ($(bot) + (-4,0)$) -- ($(bot) + (4,0)$);
\draw[very thick] ($(top) + (-4,0)$) -- ($(top) + (4,0)$);

% Coord
\node (x) at ($(bot) + (5,0)$) {$\bs{x}$};
\node (y) at ($(bot) + (0,5)$) {$\bs{y}$};
\draw[-stealth] (bot)  -- (x);
\draw[-stealth] ($(bot) + (0,-0.01)$)  -- (y);

% Nodes
\node (n0) at ($(top) + (-2,-0.4*0)$) {};
\node[draw,circle,scale=0.25*1.5] (n1) at ($(top) + (0,-0.4*1)$) {};
\node[draw,circle,scale=0.25*1.5] (n2) at ($(top) + (0,-0.4*2)$) {};
\node[draw,circle,scale=0.25*1.5] (n3) at ($(top) + (0,-0.4*3)$) {};
\node[draw,circle,scale=0.25*1.5] (n4) at ($(top) + (0,-0.4*4)$) {};
\node[draw,circle,scale=0.25*1.5] (n5) at ($(top) + (0,-0.4*5)$) {};
\node[draw,circle,scale=0.25*1.5] (n6) at ($(top) + (0,-0.4*6)$) {};
\node[draw,circle,scale=0.25*1.5] (n7) at ($(top) + (0,-0.4*7)$) {};
\node[draw,circle,scale=0.25*1.5] (n8) at ($(top) + (0,-0.4*8)$) {};
\node[draw,circle,scale=0.25*1.5] (n9) at ($(top) + (0,-0.4*9)$) {};
\node (n10) at ($(top) + (0,-0.4*10)$) {};

%\draw[dotted] (n0) -- (n1);
%\draw[dotted] (n1) -- (n2);
%\draw[dotted] (n2) -- (n3);
%\draw[dotted] (n3) -- (n4);
%\draw[dotted] (n4) -- (n5);
%\draw[dotted] (n5) -- (n6);
%\draw[dotted] (n6) -- (n7);
%\draw[dotted] (n7) -- (n8);
%\draw[dotted] (n8) -- (n9);
%\draw[dotted] (n9) -- (n10);

% Extra
\draw[>=latex, <->] ($(top) + (-3,0)$) -- ($(bot) + (-3,0)$);
\node[draw=none] (num) at ($(n5) + (0.5,0)$) {$N$};
\node[draw=none] (num) at ($(top) + (-3.5,-2)$) {$H$};

\draw[-stealth] ($(bot) + (1,0.25)$) -- ($(bot) + (2,0.25)$) node[above] {$\bs{V}$};

\end{tikzpicture}